\documentclass[]{scrartcl}

\usepackage{bibentry}
\usepackage[
breaklinks=true,
pagebackref=true,
colorlinks=false,
citecolor=SchoolColor,
filecolor=black,
linkcolor=black,
urlcolor=red]{hyperref}
\usepackage{doi}

\usepackage{xcolor}
\usepackage{mathtools}
\usepackage{amssymb}
\definecolor{PastelRed}{rgb}{1.0, 0.41, 0.38}
\usepackage{mathcommands}
\usepackage{array}
\usepackage{esdiff}
\usepackage{graphicx}
\graphicspath{ {./figures/} }
\usepackage[ruled,vlined]{algorithm2e}
\SetKwInput{KwInput}{Input}                % Set the Input
\SetKwInput{KwOutput}{Output}              % set the Output
\usepackage{multirow}
\usepackage{longtable}

\usepackage{float}
\usepackage{breakcites}

\usepackage{booktabs}
\usepackage{makecell}

\definecolor{blue}{RGB}{38,139,210}
\definecolor{cyan}{RGB}{42,161,152}
\definecolor{violet}{RGB}{108,113,196}
\definecolor{red}{RGB}{220,50,47}
\definecolor{base01}{RGB}{88,110,117}
\definecolor{base02}{RGB}{7,54,66}
\definecolor{base03}{RGB}{0,43,54}

\definecolor{OrangeCNNNeuron}{RGB}{180,74,16}
\definecolor{MagentaCNNNeuron}{RGB}{40,12,112}

\newcommand{\keywords}[1]{%
	\vspace{0.5cm}\noindent\textbf{Keywords:} #1
	\vspace{0.5cm}
}

\title{Deep Learning with CNNs: A Compact Holistic Tutorial with Focus on Supervised Regression (Preprint)}
\author{
Yansel Gonzalez Tejeda and Helmut A. Mayer \\\\
	Department of Artificial Intelligence and Human Interfaces\\
	Paris Lodron University of Salzburg
}

\begin{document}

\maketitle

\begin{abstract}
In this tutorial, we present a compact and holistic discussion of Deep Learning with a focus on Convolutional Neural Networks (CNNs) and supervised regression. While there are numerous books and articles on the individual topics we cover, comprehensive and detailed tutorials that address Deep Learning from a foundational yet rigorous and accessible perspective are rare. Most resources on CNNs are either too advanced, focusing on cutting-edge architectures, or too narrow, addressing only specific applications like image classification.This tutorial not only summarizes the most relevant concepts but also provides an in-depth exploration of each, offering a complete yet agile set of ideas. Moreover, we highlight the powerful synergy between learning theory, statistic, and machine learning, which together underpin the Deep Learning and CNN frameworks. We aim for this tutorial to serve as an optimal resource for students, professors, and anyone interested in understanding the foundations of Deep Learning. Upon acceptance we will provide an accompanying repository under \href{https://github.com/neoglez/deep-learning-tutorial}{https://github.com/neoglez/deep-learning-tutorial}
\end{abstract}


\section{Introduction}
In this tutorial, we discuss the theoretical foundations of \textbf{Artificial Neural Networks} (ANN) in its variant with several intermediate layers, namely Deep Artificial Neural Networks. More specifically, we focus on a particular ANN known as \textbf{Convolutional Neural Network} (CNN).

In order to introduce the 
methods and artifacts we will use, we \textit{mostly} follow the classical 
literature on 
\textbf{Artificial 
	Intelligence} (AI) from an overall perspective \cite{aima.2010} and \textbf{Deep 
	Learning} (DL) \cite{Goodfellow.2016}. We begin by motivating DL in Sec. \ref{sec:dl_motivation} and enunciating its standard taxonomy. Since ANNs are learning systems, we first 
expose \textbf{machine learning} (ML) concepts in Sec. \ref{sec:ml}. We then discuss the general structure of an ANN (Sec. \ref{sec:ann}) and the \textbf{concept of depth} in that context.
Convolutional Neural 
Networks are named as such because \textbf{convolutions} are a key 
building block of their \textbf{architecture}. For that reason, in Sec. \ref{sec:cnn}, we incorporate the 
convolution operation to complete our exposition of CNNs. Finally, in Sec. \ref{sec:summary}, we summarize the more relevant facts.

The theoretical deep learning framework is immense and has many ramifications 
to 
learning theory, probability and statistics, optimization, and even cognitive 
theory, to 
mention just a few 
examples. There are several books for every topic described in the subsections of 
this tutorial, not to mention articles in conferences and journals. Therefore, our presentation of deep learning here is highly 
selective and aims to expose the fundamental aspects needed to agile assimilate DL. 

It is worth noting that, since DL is a \textit{relative} young field, the 
terminology 
greatly varies. We often indicate when several terms are used to designate a 
concept, but the reader should be aware that this indication is by far not 
exhaustive. Notation is at the end of this manuscript.

We begin by motivating the DL framework in Sec. \ref{sec:dl_motivation}. As its name suggests, DL is a subset of \textbf{Machine Learning} (ML); for that reason, in Sec. \ref{sec:ml}

Then, in Sec. \ focusing the discussion in a pivotal element, namely the dataset. Without a dataset is  

%Interestingly, the fundamental methods of DL have kept invariable the laust 
%decade.

\section{Motivation}\label{sec:dl_motivation}

Intelligent agents need to learn in order to acquire knowledge from their environment. To perform various tasks, humans learn, among other things, from examples. In this context, learning means that for a defined task, the agent must be able to generalize. That is, the agent should successfully perform the task beyond the examples it has seen before, and typically, we expect the agent to make \textit{some form} of prediction. Most tasks the agent performs can be framed as either classification or regression. For both types of tasks, when the examples are accompanied by known \textbf{target values} or \textbf{labels}, which the agent aims to predict, we refer to this as \textbf{supervised learning}, as a \textbf{learning paradigm}. From a \textit{pedagogical perspective}, the agent is being supervised by receiving the \textbf{ground truth} corresponding to each learning example.

Besides supervised learning, there are three other learning paradigms: 
\textbf{unsupervised learning}, \textbf{reinforcement leaning} and 
\textbf{semi-supervised learning}. When no 
targets are available, or not provided, learning must occur in an 
unsupervised manner. If, alternatively, the targets are noisy or a number of 
them is missing, learning must proceed semi-supervised. In contrast, if no ground 
truth is provided at all, but a series of reinforcements can be imposed, the agent 
can nonetheless learn by being positively or negatively (punished) rewarded. Here, we focus on supervised learning.

\section{Machine Learning}\label{sec:ml}

From a general perspective the learning agent has access to a set of ordered 
examples

\begin{equation}\label{eq:examples}
	\mathbb{X} \stackrel{\text{def}}{=} \{\boldsymbol{x}^{(1)}, \cdots , 
	\boldsymbol{x}^{(m)}\}
\end{equation}

assumed to have been drawn from some 
unknown 
generating 
distribution $p_{data}$. A learning example $\boldsymbol{x}^{(i)}$ (also a 
\textbf{pattern}) may be 
\textit{anything} that 
can be represented in a computer, for example, a scalar, a text excerpt, an image, 
or a song. 
Without loss of generality, we will assume that every $\boldsymbol{x}^{(i)}$ 
is a vector 
$\boldsymbol{x}^{(i)} \in \mathbb{R}^n$, and its components the example 
\textbf{features}. For instance, one could represent an image as a vector by 
arranging its pixels to form a sequence. Similarly, a song could be vectorized 
by forming a sequence of its notes. We will clarify when we refer to a 
different 
representation.

Furthermore, the examples are endowed with their 
\textbf{ground truth} 
(set of \textbf{labels}, also \textbf{supervision signal}) $\mathbb{Y}$ so that for every example 
$\boldsymbol{x}^{(i)}$, 
there exists a 
label (possibly a vector) $\boldsymbol{y}_i \in 
\mathbb{Y}$. Using these examples the agent can learn to perform essentially two 
types of tasks: \textbf{regression} and \textbf{classification} (or both). 
Regression may be defined as estimating \textbf{one or more values} $y \in 
\mathbb{R}$, for 
instance, human body dimensions like height or waist circumference. In a classification scenario, the agent is required to assign 
\textbf{one or more of a finite set of categories} to 
every input, for example, 
from an 
image of a person, designate which body parts are arms or legs.

Additionally, in \textbf{single-task learning} the agent must learn to predict a 
single 
quantity (binary or multi-class classification, univariate regression), while in 
\textbf{multi-task learning} the algorithm must learn to estimate several 
quantities, e.g., 
multivariate multiple regression.

\subsection{Learning Theory}\label{subsec:learning_theory}

In general, we consider a hypothesis space $\mathcal{H}$, consisting of 
functions. For example, one could consider the space of linear functions. 
Within $\mathcal{H}$, the function

\begin{equation}\label{eq:real_model}
	f(\mathbb{X}) = \mathbb{Y}
\end{equation} maps the examples (inputs) to their labels (outputs). The learning 
agent can be then expressed as a model 

\begin{equation}\label{eq:approx_model}
	\hat{f}(\mathbb{X}) = \hat{\mathbb{Y}}
\end{equation} that approximates $f$, where $\hat{\mathbb{Y}}$ is the model 
estimate of the targets $\mathbb{Y}$.

Now, the ultimate goal of the learning agent is to perform well beyond the 
examples it has seen before, i.e., to exhibit \textbf{good generalization}. More 
concretely, learning must have low \textbf{generalization error}. To 
assess generalization, the \textbf{stationarity assumption} must be adopted to 
connect the observed examples to the hypothetical future examples that the agent 
will perceive. This assumption postulates that the examples are sampled from a 
probability distribution $p_{data}$ that remains stationary over time. 
Additionally, it is assumed that a) each example $\boldsymbol{x}^{(i)}$ is 
independent of the 
previous examples and b) that all examples have identical prior probability 
distribution, abbreviated \textbf{i.i.d.}.

With this in hand, we can confidently consider the observed examples as a 
\textbf{training set} 
$\{x_{train}, y_{train}\}$ (also \textbf{training data}) and the 
future examples as a \textbf{test set} $\{x_{test}, y_{test}\}$, where 
generalization will be assessed. As the data 
generating distribution $p_{data}$ is assumed to 
be unknown (we will delve into this in subsection \ref{subsec:stats}), the 
agent has only access to the training set. One intuitive strategy for the learning 
agent to estimate generalization error is to minimize the \textbf{training error} 
computed on the train set, also termed \textbf{empirical error} or 
\textbf{empirical risk}. Thus, this strategy is called \textbf{Empirical Risk 
	Minimization} 
(ERM). Similarly, the \textbf{test error} is calculated on the test set. Both 
training and test errors are defined in a general form as 
the sum of incorrect estimated targets by the model (in Subsec. 
\ref{subsubsec:loss_function} we restate this definition), for 
example, in the case of 
$m$ training examples, the training 
error $E_{train}$ is

\begin{equation}\label{eq:train_test_error_def}
	E_{train} = \sum_{i = 1}^{m} \big( f^*(\boldsymbol{x}^{(i)}) \neq 
	\boldsymbol{y}_i \big).
\end{equation}

Note that given a specific model $f^*$, and based on the i.i.d. assumption the 
expected training error equals the expected test error. In practice, the previous 
is greater than or equal to the latter. Here, the model may exhibit two important 
flaws: \textbf{underfitting} or \textbf{overfitting}. A model that underfits 
does not achieve a small training error (it is said that it can not capture 
the underlying structure of the data). Overfitting is the contrary of 
underfitting, it 
occurs when the model ``memorizes" the training data, ``perhaps like the everyday 
experience that a person who provides
a perfect detailed explanation for each of his single actions may raise 
suspicion" \cite{Shalev-Shwartz2014-zp}. When evaluated, a model that 
underfits yields a high difference between the training and the test error.

\subsection{Model Evaluation}

Assessing generalization by randomly splitting the available data in a training 
set and a test set is called \textbf{holdout cross-validation} because the test 
set is kept strictly separate from the training set and used only once to report 
the algorithm results. However, this model evaluation technique has two important 
disadvantages:

\begin{itemize}
	\item It does not use all the data at hand for training, a problem that is 
	specially relevant for small datasets.
	\item In practice, there can be the case where the i.i.d. assumption does not 
	hold. Therefore, the assessment is highly sensitive to the training/test split.
\end{itemize}

A remedy to these problems is the $k$\textbf{-fold cross-validation} evaluation 
method (cf. \cite{hastie_09_elements-of.statistical-learning} p. 241 - Cross 
Validation). It splits the available 
data into $k$ nonoverlapping 
subsets of equal size. Recall the dataset has $m$ examples. First, learning is 
performed $k$ times with $k-1$ subsets. Second, in every iteration, the test 
error is 
calculated on the subset that was not used for training. Finally, the model 
performance is reported as the 
average of the $k$ test errors. The cost of using this method is the 
computational overload, since training and testing errors must be computed $k$ 
times.

\subsection{Statistics and Probability Theory}\label{subsec:stats}

To further describe machine learning we borrow statistical concepts. In order to 
be consistent with the established literature\cite{Goodfellow.2016}, 
we use in this subsection $\boldsymbol{\theta}$ and $\boldsymbol{\hat{\theta}}$ to 
to denote a quantity and its estimator.

The 
agent must learn to estimate the wanted quantity $\boldsymbol{\theta}$, say. From 
a statistical 
perspective the agent performs a point estimate $\boldsymbol{\hat{\theta}}$. In 
this view the 
data points are the i.i.d. learning examples $\{\boldsymbol{x}^{(1)}, \cdots , 
\boldsymbol{x}^{(m)}\}$. A 
point estimator is a 
function of the data such as $\boldsymbol{\hat{\theta}}_m \stackrel{\text{def}}{=} 
g(\{\boldsymbol{x}^{(1)}, \cdots , 
\boldsymbol{x}^{(m)}\})$, with $g(x)$ being a very general suitable function. Note 
that we can 
connect the quantity $\boldsymbol{\theta}$ and its estimator 
$\boldsymbol{\hat{\theta}}$ to the functions $f$ and $\hat{f}$ in Equations 
\ref{eq:real_model} and \ref{eq:approx_model}. Given the 
hypothesis space $\mathcal{H}$ that embodies possible input-output relations, the 
function $\hat{f}$ that approximates $f$ can be treated as a point estimator in 
function space.

Important properties of estimators are bias, variance and standard error. The 
estimator bias is a measure of how much the real and the estimated value 
differ. It 
is defined as $bias(\boldsymbol{\hat{\theta}}_m) = 
\mathbb{E}(\boldsymbol{\hat{\theta}}_m) - \boldsymbol{\theta}$. Here 
the 
expectation $\mathbb{E}$ is taken over the data points (examples).

An unbiased 
estimator has $bias(\boldsymbol{\hat{\theta}}_m) = 0$, this implies that 
$\boldsymbol{\hat{\theta}}_m = 
\boldsymbol{\theta}$. An asymptotically unbiased estimator is when $\lim_{m \to 
	\infty} 
bias(\boldsymbol{\hat{\theta}}_m) = 0$, which implies that $\lim_{m \to \infty} 
\boldsymbol{\hat{\theta}}_m = \theta$. An example of unbiased estimator is the 
mean estimator 
(mean of the training examples) of the Bernoulli distribution with mean 
$\boldsymbol{\theta}$.

The variance $Var(\boldsymbol{\hat{\theta}}) = $ and 
Standard 
Error 
$SE(\boldsymbol{\hat{\theta}})$ of the 
estimator are useful for comparing different experiments. A good estimator has low 
bias and low variance.

Let us now discuss two important estimators: the Maximum Likelihood and the 
Maximum a Posteriori estimators.

\subsubsection{Maximum Likelihood Estimation}

To guide the search for a good estimator $\boldsymbol{\hat{\theta}}$, the 
$\textbf{frequentist approach}$ is usually adopted. The wanted quantity 
$\boldsymbol{\theta}$, possibly multidimensional, is seen as fixed but unknown, 
and the observed data is 
random. The parameters $\boldsymbol{\theta}$ govern the data generating 
distribution $p_{data}(x)$ from which the observed i.i.d. data 
$\{\boldsymbol{x}^{(1)}, 
\cdots , \boldsymbol{x}^{(m)}\}$ arose.

Then, a parametric model for the observed data is presumed, i.e., a probability 
distribution $p_{model}(\mathbf{x}; \boldsymbol{\theta})$.  For 
example, if the 
observed data is presumed to have a normal distribution, then the parameters are 
$\boldsymbol{\theta} 
= \{\mu, \sigma\}$, and the model $p_{model}(x; \boldsymbol{\theta}) = 
\frac{1}{\sigma\sqrt{2\pi}} e^{-\frac{1}{2}\left(\frac{x - 
		\mu}{\sigma}\right)^2}$. Here candidates parameters
$\boldsymbol{\theta}$ must be considered, and ideally, $p_{model}(\mathbf{x}; 
\boldsymbol{\theta}) \approx p_{data}(\mathbf{x})$. Next, a $\textbf{likelihood 
	function}$ 
$L$ 
can be defined as the mapping between the data $\mathbf{x}$ and a given 
$\boldsymbol{\theta}$, to a real number estimating the true probability 
$p_{data}(\mathbf{x})$. Since the observations are assumed to be independent, 
\begin{equation}
	L(\mathbf{x}; \boldsymbol{\theta}) = \prod_{i=1}^{m} p_{model}( 
	\boldsymbol{x}^{(i)}, 
	\boldsymbol{\theta}).
\end{equation}

The \textbf{Maximum Likelihood (ML) estimator} is the estimator 
$\boldsymbol{\hat{\theta}}$ 
that, among all candidates, chooses the parameters that make the observed data 
most probable.

\begin{align}\label{eq:max_likelihood_est}
	\boldsymbol{\hat{\theta}}_{ML}& = \argmax_{\boldsymbol{\theta}} L(\mathbf{x}; 
	\boldsymbol{\theta}) \\
	&=\label{eq:max_likelihood_est_prod} \argmax_{\boldsymbol{\theta}} 
	\prod_{i=1}^{m} p_{model}(\boldsymbol{x}^{(i)}, 	\boldsymbol{\theta})
\end{align}

Taking the logarithm of the right side 
in \ref{eq:max_likelihood_est_prod} facilitates the probabilities computation and 
does 
not change the maximum location. This leads to the $\textbf{Log-Likelihood}$

\begin{align}\label{eq:max_log_likelihood}
	\boldsymbol{\hat{\theta}}_{ML} = \argmax_{\boldsymbol{\theta}} \sum_{i=1}^{m} 
	\log \, p_{model}(\boldsymbol{x}^{(i)}, \boldsymbol{\theta})
\end{align}

An insightful connection to \textbf{information theory} can be established by 
transforming 
Eq. \ref{eq:max_log_likelihood}. Firstly, dividing the right term by the constant 
$m$ does not shift the maximum in the left term, i.e.,

\begin{align}\label{eq:max_log_likelihood_expec}
	\boldsymbol{\hat{\theta}}_{ML} = \argmax_{\boldsymbol{\theta}} \frac{1}{m}  
	\sum_{i=1}^{m} \log \, p_{model}(\boldsymbol{x}^{(i)}, \boldsymbol{\theta}).
\end{align}

Secondly, by definition, the empirical mean $ \frac{1}{m}  
\sum_{i=1}^{m} \log \, p_{model}(\boldsymbol{x}^{(i)}, \boldsymbol{\theta})$ of 
the assumed 
model distribution $p_{model}$ 
equals the \textbf{expectation given its empirical distribution} 
$\mathbb{E}_{\textbf{X} 
	\sim \hat{p}_{data}}$ defined by the training 
set $\hat{p}_{data}(x)$. A discussion of the empirical distribution 
$\hat{p}_{data}(x)$ is 
out of the scope, but it suffices to say that it is a method to approximate the 
real underlying distribution of the training set (not to be confused with the 
data-generating distribution $p_{data}$). Finally, Eq. 
\ref{eq:max_log_likelihood_expec} can be written as 

\begin{align}\label{eq:max_log_likelihood_expec_complete}
	\boldsymbol{\hat{\theta}}_{ML} = \argmax_{\boldsymbol{\theta}} 
	\mathbb{E}_{\textbf{X} 
		\sim \hat{p}_{data}} \log \, p_{model}(\boldsymbol{x}, 
	\boldsymbol{\theta}).
\end{align}

Information theory allows to evaluate the degree of dissimilarity between the 
empirical distribution of the 
training data $\hat{p}_{data}(x)$ and the assumed model distribution 
$p_{model}(x)$ using the Kullback–Leibler divergence $D_{KL}$ 
(also called relative 
entropy) as 

\begin{equation}\label{eq:kl_divergence}
	D_{KL}(\hat{p}_{data} \| p_{model}) = \mathbb{E}_{\textbf{x} \backsim 
		\hat{p}_{data}} [\log \hat{p}_{data}(\boldsymbol{x}) - \log 
	p_{model}(\boldsymbol{x})].
\end{equation}

Here the expectation $\mathbb{E}$ is taken over the training set because 
Eq.\ref{eq:kl_divergence} quantifies the additional uncertainty arising from using 
the assumed model $p_{model}$ to predict the training set $\hat{p}_{data}$. 

Because we want the divergence between these two distribution to be as small as 
possible, it makes sense to minimize $D_{KL}$. Note that the term 
$\log \, \hat{p}_{data}(\boldsymbol{x})$ in Eq. \ref{eq:kl_divergence} can not be 
influenced by optimization, since it 
has been completely determined by the data generating distribution $p_{data}$. 
Denote the minimum $D_{KL}$ as $\boldsymbol{\theta}_{D_{KL}}$, then

\begin{equation}
	\boldsymbol{\theta}_{D_{KL}} = 
	\argmin_{\boldsymbol{\theta}} -  \mathbb{E}_{\textbf{x} \backsim 
		\hat{p}_{data}} [\log p_{model}(\boldsymbol{x}, \boldsymbol{\theta})]
\end{equation}

The expression inside the arg min is \textbf{the cross-entropy} of 
$\hat{p}_{data}$ and $p_{model}$. Comparing this with Eq. 
\ref{eq:max_log_likelihood_expec_complete} it can be arrived to the conclusion 
that 
maximizing the likelihood coincides exactly with minimizing the cross-entropy 
between the distributions.  

%To adjust \ref{eq:max_likelihood_est} to supervised learning, the conditional 
%probability $P(\textbf{y} | \textbf{x}; \boldsymbol{\theta})$ is used to predict 
%the labels given the input. Additionally, taking the logarithm of the right side 
%in \ref{eq:max_likelihood_est} facilitates the probabilities computation and does 
%not change the maximum location. These two transformations allows using the 
%$\textbf{Log-Likelihood}$:

\subsubsection{Maximum a Posteriori Estimation}

In contrast to the frequestist approach, the \textbf{Bayesian approach} considers 
$\boldsymbol{\theta}$ to be a random variable and the dataset $\textbf{x}$ to be 
fixed and directly observed. A \textbf{prior probability} distribution  
$p(\boldsymbol{\theta})$ must be defined to express the degree of uncertainty of 
the state of knowledge before observing the data. In the absence of more 
information, a Gaussian distribution is commonly used. The Gaussian distribution 
is ``the least surprising and least informative assumption to make" 
\cite{mcelreath2020statistical}, i.e., it is the distribution with maximum entropy 
of all. Therefore, the Gaussian is an appropriate starting prior when 
conducting Bayesian inference.

After observing the data, Bayes' rule can be applied to find \textbf{the posterior 
	distribution} of the parameters given the data $p(\boldsymbol{\theta} | 
\mathbf{x}) 
= \frac{p(\mathbf{x} | 
	\boldsymbol{\theta})p(\boldsymbol{\theta})}{p(\mathbf{x})}$, where 
$p(\mathbf{x} | \boldsymbol{\theta})$ is the likelihood and 
$p(\mathbf{x})$ the probability of the data.

Then, the \textbf{Maximum a Posteriori (MAP) estimator} selects the point of 
maximal posterior probability:

\begin{equation}\label{eq:map_estimator}
	\boldsymbol{\hat{\theta}}_{MAP} = \argmax_{\boldsymbol{\theta}} 
	p(\boldsymbol{\theta} | \mathbf{x}) = \argmax_{\boldsymbol{\theta}} log \,  
	p(\textbf{x} | \boldsymbol{\theta}) + log \, p(\boldsymbol{\theta}).
\end{equation}

\subsection{Optimization and Regularization}

\subsubsection{Loss Function}\label{subsubsec:loss_function}

The training error as defined by Eq. \ref{eq:train_test_error_def} is an 
\textbf{objective function} (also \textbf{criterion}) that needs to be minimized. 
Utility theory regards this objective as a 
\textbf{loss function} $J(x): \mathbb{R}^n \to \mathbb{R}$ because the agent 
incurs in a lost of expected 
utility 
when 
estimating the wrong output with respect to the expected utility when estimating 
the correct value. To 
establish the difference of these values, a determined distance function must be 
chosen.

For general (multivariate, multi-target) regression tasks the \textbf{p-norm} is 
preferred. The \textbf{p-norm} is calculated on the vector of the corresponding 
train 
or test error ${\Vert \hat{\boldsymbol{y}}_i - \boldsymbol{y}_i \Vert}_p = \left( 
\sum_{i} | \hat{\boldsymbol{y}}_i - \boldsymbol{y}_i |^p \right)^\frac{1}{p}$, 
which for $p = 2$ reduces to the Euclidean norm. 
Since the \textbf{Mean Squared Error} (MSE) ${\Vert \hat{\boldsymbol{y}}_i - 
	\boldsymbol{y}_i \Vert}_2^2$ can be faster computed,  it is 
often used for optimization. We used this loss function (also \textbf{cost 
	function}) when training our 
network (\ref{fig:neural_anthro}) with $m$ examples:

\begin{align}\label{eq:mse}
	J(y) &= \frac{1}{m} \sum_{i = 1}^{m} \left( 
	\hat{\boldsymbol{y}}_i - \boldsymbol{y}_i \right)^2\\
	&= \frac{1}{m} \sum_{i = 1}^{m} \frac{1}{2} \sum_{j = i}^{|\boldsymbol{y}|} 
	\left( 
	\hat{y}_j - y_j \right)^2.
\end{align}

Since we regard every target $\boldsymbol{y}_i$ as a vector, the inner sum runs 
over its components $y_i$ and $|\boldsymbol{y}|$ is its length. The rather 
arbitrary scaling factor $2$ in the denominator is a mathematical convenience 
for the discussion in \ref{subsec:training_ann}.

For classification tasks the \textbf{cross-entropy loss} is commonly adopted.

\subsubsection{Gradient-Based Learning}\label{subsubsec:gradient_descent}

As we will see in Sec. \ref{sec:ann}, in the context of Deep Learning, the  
parameters $\boldsymbol{\theta}$ discussed in Subsec. \ref{subsec:stats} are named 
\textbf{weights} $\boldsymbol{w}$. Without loss of generalization, we will 
consider the parameters to be a \textit{collection} of weights. We do not lose 
generalization because the discussion here does not depend on the specific form 
of the cost function.

Therefore, 
parameterizing the estimator 
function by the 
weights, minimization of the cost function can be written as

\begin{equation}\label{eq:min_loss}
	J^*(\boldsymbol{w};  \boldsymbol{x}, y) =  \argmin_{\boldsymbol{w}} 
	\frac{1}{m} \sum_{i = 1}^{m} 
	\left( 
	\hat{f}( \boldsymbol{x}^{(i)}, \boldsymbol{w}) - y_i \right)^2
\end{equation}

The function $\hat{f}( \boldsymbol{x}^{(i)}, \boldsymbol{w})$ is, in a broader 
sense, a 
nonlinear model, 
thus the minimum of the loss function has no closed form and numeric 
optimization must be employed. Specifically, $J$ is iteratively decreased in the 
orientation of the negative gradient, until a local or global minimum is reached, 
or the optimization converges. This optimization method (also 
\textbf{optimizer}) is termed \textbf{gradient descent}. Because optimizing in the 
context of machine learning may be interpreted as learning, a 
\textbf{learning 
	rate} $\epsilon$ is introduced to influence the step size at each iteration. 
The 
gradient descent optimizer changes 
weights using the 
gradient $\nabla_w$ of $J(\boldsymbol{w})$ according to the \textbf{weight update 
	rule}

\begin{equation}\label{eq:gradient_descent}
	\boldsymbol{w} \leftarrow \boldsymbol{w} - \epsilon \nabla_w 
	J(\boldsymbol{w}) 
\end{equation}

We postpone the discussion on Stochastic Gradients Descent (SGD) to insert it in 
the context of training neural networks (Subsec. \ref{subsec:training_ann}).

\subsubsection{Regularization}\label{subsec:regularization}

The essential problem in machine learning is to design algorithms that 
perform well both in training and test data. However, because the true 
data-generating process is unknown, merely decreasing the training 
error does not guarantee a small generalization error. Even worst, extremely 
small training errors may be a symptom of overfitting.

\textbf{Regularization} intends to improve generalization by combating 
overfitting. It 
focuses on reducing test error using constraints and penalties on the model family 
being trained, which can be interpreted as pursuing a more \textbf{regular} 
function.

One such an important penalty may be incorporated into the objective function. 
More specifically, the models parameters may be restricted by a function 
$\Omega(\boldsymbol{w})$ to obtain the modified cost function 
$\tilde{J}(\boldsymbol{w})$ with two terms as

\begin{equation}\label{eq:min_regularized_loss}
	\tilde{J}(\boldsymbol{w}) =  \argmin_{\boldsymbol{w}} \, \frac{1}{m} 
	\sum_{i = 1}^{m} 
	\left( 
	\hat{f}(\boldsymbol{x}^{(i)}, \boldsymbol{w}) - y_i \right)^2 + \, 
	\alpha \Omega(\boldsymbol{w}).
\end{equation}

The second term's 
coefficient $\alpha \in [0, \infty)$ ponders the penalty's influence relative to 
$J^*$. The specific form of $\Omega(\boldsymbol{w})$ (also \textbf{regularizer}) 
is usually a $L^1 = {\Vert 
	\boldsymbol{w} \Vert}_1$ or $L^2 = {\Vert \boldsymbol{w} \Vert}_2^2$ 
\textbf{norm penalty}. Choosing one of these norm penalties expresses a preference 
for a determined function class. For example, the $L^1$ penalty enforces 
\textbf{sparsity} on the weights, i.e., pushes the weights towards zero, while the 
effect of $L^2$ norm causes the weights of less influential features to decay 
away. Because of 
this the $L^2$ penalty norm is called \textbf{weight decay}.

Weight decay has lost popularity in favor of other 
regularization techniques (Subsec. \ref{subsec:batch_norm}). It worth saying, 
however, that weight decay is an 
uncomplicated effective 
method for enhancing generalization \cite{NIPS1991_weight_decay}. Modifying Eq. 
\ref{eq:min_loss} to 
include the penalty with $\alpha = \frac{1}{2}$, the cost function becomes

\begin{equation}\label{eq:min_l2_regularized_loss}
	\tilde{J}(\boldsymbol{w}) =  \argmin_{\boldsymbol{w}} \, \frac{1}{m} 
	\sum_{i = 1}^{m} 
	\left( 
	\hat{f}(\boldsymbol{x}^{(i)}, \boldsymbol{w}) - y_i \right)^2 + \, 
	\frac{\alpha}{2} \boldsymbol{w}^T \boldsymbol{w}.
\end{equation}

Correspondingly, after substituting and reorganizing terms, the weight update rule in Eq. \ref{eq:gradient_descent} results in

\begin{equation}\label{eq:gradient_descent_regularized}
	\boldsymbol{w} \leftarrow (1 - \epsilon \alpha)\boldsymbol{w} - \epsilon 
	\nabla_w J(\boldsymbol{w}).
\end{equation}

\section{Deep Forward Artificial Neural Networks}\label{sec:ann}

Deep Forward Artificial Neural Networks (DFANN) is a model conveying several 
important interlaced concepts. We start with its building blocks and then connect 
them to form the network. Additionally, to describe the parts of an ANN, the term 
\textbf{architecture} is used.

\subsection{Artificial Neural Networks}

\textbf{Artificial Neural Networks} (ANNs) is a computational model originally 
inspired by the functioning of the human brain \cite{McCulloch1943, 
	hebb2005organization, Rosenblatt_1957_6098}. The model exploits the structure of 
the brain, 
focusing on neural cells 
(\textbf{neurons}) and its 
interactions. These neurons have three functionally distinct parts: 
\textbf{dendrites}, 
\textbf{soma} and \textbf{axon}. The dendrites gather information from other 
neurons and transmit it to the soma. Then, the soma performs a relevant nonlinear 
operation. If 
exceeding a determined threshold, the result causes the neuron to emit an ouput 
signal. 
Following, the output signal is delivered by the axon to others neurons. 

\subsubsection{Artificial Neurons}

Similarly, the elementary \textbf{processing unit} of ANNs is the 
\textbf{artificial neuron}. The 
neurons receive a number of inputs with \textbf{associated intensity} (also 
strength) and combine these 
inputs in a nonlinear manner to 
produce a determined output that is transmitted to other neurons. Depending on the 
form of the non-linearity, it can be said that the neuron `fires' if a determined 
threshold is reached.

Every input to the neuron has an associated scalar
called \textbf{weight} $w$. The non-linearity, also called \textbf{activation 
	function} is 
denoted by $g(\cdot)$. The neuron first calculates an affine transformation of its 
inputs $z = 
\sum_{i=1}^{n} w_i \cdot x_i + c$. The intercept (also \textbf{bias}) $c$ may be 
combined with a \textit{dummy input} $x_{dummy} = 1$ to simplify notation 
such as the affine transformation can be written in matrix form $\sum_{i=0}^{n} 
w_i \cdot x_i = \textbf{w} \cdot \textbf{x}$. One of the reason ANNs where 
invented is to overcome the limitations of linear models. Since a linear 
combination of linear units is 
also linear, a non-linearity is needed to increase the capacity of the model. 
Accordingly, the neuron applies the activation function to 
compute the \textbf{activation} $a = g(\textbf{w} \cdot 
\textbf{x})$, 
which is 
transmitted 
to other neurons.

Referring to activation functions, \cite{Goodfellow.2016} states 
categorically ``In 
modern neural networks, the default recommendation is to use the \textbf{rectified 
	linear unit} or ReLU" introduced by \cite{jarrett2009best}:

\begin{equation}\label{eq:relu}
	g(z) = max(0, z).
\end{equation}

In 
order to enable learning with gradient 
descent (Eq. \ref{eq:gradient_descent}) as described 
in Subsec. 
\ref{subsubsec:gradient_descent}, is desirable the activation function to be 
continuous differentiable.
Although the ReLU is not differentiable at $z = 0$, in practice, this 
does not constitute a problem. Practitioners assume a `derivative default value' 
$f'(0) = 0$, $f'(0) = 1$ or values in between.

Other activation functions are the \textbf{logistic sigmoid} $\sigma(z) = 1 /( 1 + 
exp(-z))$ and the \textbf{softplus function} $\zeta(z) = \log (1 + 
exp(z))$. As a side note, a neuron must not 
necessarily compute a 
non-linearity, and in that 
case, it is called a \textbf{linear unit}.

The \textbf{perceptron} is an artificial neuron introduced by 
\cite{Rosenblatt_1957_6098}, who demonstrated that only one neuron suffices to 
implement a \textbf{binary classifier}. The perceptron had originally a hard 
threshold activation function, but it can be generalized to use any of the 
above-mentioned activation function. For example, if the perceptron uses a 
logistic sigmoid activation function, then the output can be interpreted as the 
probability of some event happening, e.g., \textbf{logistic regression}.

\subsubsection{Deep Feedforward Networks}\label{subsubsec:deep_forward_networks}

Connecting the artificial neurons results in an artificial neural network with at 
least two distinguishable \textbf{layers}: the \textbf{input layer} and the 
\textbf{output layer}. The input layer is composed by the units that directly 
accept the features (or their preprocessed representation) of the learning 
examples. The output layer contains the units that produce 
the answer to the learning task.

A \textbf{single-layer neural network} (sometimes called single-layer perceptron 
network, or simply perceptron network) (SLNN) is the most basic structure 
that can be assembled by directly 
connecting any number of input units to any number of output units. Note that, by 
convention, the input layer is not counted. Despite this 
type of network being far 
from useless, they are limited in their expressive power. Perceptron networks are 
unable to represent a \textbf{decision boundary} for \textbf{non-linearly 
	separable} 
problems(\ref{minsky1969perceptrons}). A 
problem is linearly separable if the learning examples can be separated in two 
nonempty sets,
such as every set can be assigned exclusively 
to one of the two half spaces defined by a \textbf{hyperplane} in its ambient 
space. This is important because, in general, practical problems in relevant areas 
like computer vision and natural language processing are not linearly separable.

Beyond the SLNN, networks can be extremely complex. In order to make progress, one 
important restriction that may be imposed is to model the network as a 
\textbf{directed acyclic graph} (DAG) with \textbf{no shortcuts}. The neurons are 
the 
\textbf{nodes}, 
the 
connections the \textbf{directed edges} (or \textbf{links}), and the 
\textbf{weights} the 
connection 
intensity. Because 
a 
DAG does not contain cycles, the information in the ANN is said to flow in one 
direction. For that reason, this type of ANN that does not contain 
\textbf{feedback links}, are called \textbf{feedforward networks} or 
\textbf{multi-layer perceptron} (MLP). If the network 
does have feedback links then it is called a \textbf{recurrent neural network} 
(RNN). In this tutorial we focus on feedforward networks.

In equation \ref{eq:approx_model} (Sec \ref{sec:ml}), we enunciated that the goal 
of a learning agent $\hat{f}(x)$ is to approximate some function $f(x)$, and from 
a 
statistical 
perspective, the agent is a function estimator of $\boldsymbol{\theta}$ (Subsec. 
\ref{subsec:stats}). Motivated by 
the idea 
of \textbf{compositionality}, it can be assumed that this learning agent 
$\hat{f}(\boldsymbol{x})$ is precisely the composition of a number of different 
functions
\begin{equation}\label{eq:ann_composition}
	\hat{f}(\boldsymbol{x}) = f^{n}(f^{n-1}( \cdots f^{0}(\boldsymbol{x}))) 
\end{equation}
where $f^{(0)}$ and $f^{(n)}$ represent the input and output layer, respectively.

Further, starting from the input layer and ending at the output layer, every 
function $f^{(l)}$, for $l \in {1, \cdots, n-1}$ can be associated 
with a collection of DAG units that receive inputs only from the units in
preceding collections. These functions and the units group they define, are 
called 
\textbf{hidden layers} and denoted by $h(\cdot)$.

Hidden layers are named as such because 
of their relation to the training examples. The training examples determine the 
form and behavior of the input and the output 
layer. Indeed, the input layer processes the training examples and the output 
layer 
must contribute to minimize the training loss. However, the training examples do 
not rule the intermediary layer computation scheme. Instead, the algorithm must 
learn how to adapt those layers to achieve its goal.

Finally, the length of the entire function chain $f^{(1)}, \cdots, f^{(n)}$ is the 
\textbf{depth} $\ell$ of the 
network ($f^{(0)}$ is the \textit{first layer} but is not counted for depth), 
which, combined with the concepts explained 
before, confers the name to 
the computational model: deep feedforward networks. Likewise, the \textbf{width of 
	a layer} 
is the 
number of units it contains, and the \textbf{network's width} is sometimes defined 
as the 
``maximal number of nodes in a layer" \cite{Lu2017TheEP}. Additionally, the input 
layer 
is also called the first layer and the next layers are the second, the third and 
so forth.

The \textbf{architecture of the network} designates the network's depth and width, 
as well as the layers and their connections to each other (further in Subsec. \ref{subsec:conv_layer} we show an example of a network architecture). Depending on how the 
units are connected, different layer types can be assembled. 
Note that because shortcuts are not allowed, there can not be intra-layer unit 
connections. The \textbf{fully-connected layer} $h^{(l)}$ (also \textbf{dense 
	layer}, in 
opposition to a \textbf{shallow layer}) is the 
most 
basic, which all of its units connect to all units of the preceding layer 
$h^{(l - 1)}$ though a 
\textbf{weight matrix} $\boldsymbol{W}$, i.e.,

\begin{align}
	\boldsymbol{h}^{(l)} &= g^{l}(\boldsymbol{W}^l \boldsymbol{h}^{(l-1)} + 
	\boldsymbol{b}^l)\label{eq:dense_layer}\\
	\boldsymbol{h}^{(l)} &= g^{l}(\boldsymbol{W}^l 
	\boldsymbol{h}^{(l-1)}).\label{eq:dense_layer_no_bias}
\end{align}

The weight matrix $\boldsymbol{W}^l \in \mathbb{R}^{m \times n}$ and the 
\textbf{bias vector} $\boldsymbol{b}^l \in \mathbb{R}^m$, where $m$ and $n$ are 
the width of 
the current and preceding layer, respectively. Alternatively, the notation may be 
simplified by incorporating the bias vector into the matrix with a dummy input 
(Eq. \ref{eq:dense_layer_no_bias}).

In Sec. \ref{subsub:backprop} we will further discuss the notation and in 
Subsec. 
\ref{subsec:conv_layer} will see other layer types.

One of the most significant results in feedforward networks research are the 
\textbf{universal approximation theorems}. The exact form of these theorems is 
positively technical, but at their core they demonstrate that feedforward 
networks can approximate any continuous function to any desirable level of 
accuracy. There are 
basically three theorem variants: arbitrary width and bounded depth, i.e., a 
universal approximator with 
only one hidden layer containing a sufficient number of neurons
\cite{DBLP:journals/mcss/Cybenko89}, bounded 
width and arbitrary depth \cite{DBLP:journals/jat/Gripenberg03}, i.e., a 
sufficient number of layers with a bounded amount of neurons, and remarkably, 
bounded depth and width \cite{DBLP:journals/ijon/MaiorovP99}.

%After these considerations, we may specify the model as a feedforward network 
%with 
%input layer $$

%\begin{equation}
%	
%\end{equation}

\subsection{Training}\label{subsec:training_ann}

Having described the structure of an artificial neural network, and how they learn by means of optimization, we now need to perform the actual training.
The learning examples are \textit{entered} into the network through the input layer and the network weights need to be updated using gradient descent to minimize the loss.

\subsubsection{Stochastic Gradient Descent}\label{subsubsec:sgd}

As we already discussed in Subsec. \ref{subsubsec:gradient_descent}, gradient 
descent enables optimization of general loss functions. Specifically, the model 
weights are updated using the gradient of the loss function $\nabla_w 
J(\boldsymbol{w})$, which is sometimes called \textbf{pure}, 
\textbf{deterministic} or \textbf{batch 
	gradient descent}. The problem is that the computation of this gradient 
requires 
an error summation over the entire dataset $\nabla_w J(\boldsymbol{w}) =  \sum_{i 
	= 1}^{m}(\hat{y}_i - y_i)x_i$. In practice, this computation can be extremely 
expensive or 
directly intractable.

Stochastic gradient descent (SGD) proposes a radical solution to this problem. It 
does not 
compute the gradient exactly. Instead, it estimates the gradient with one training 
example $\boldsymbol{x}^{(i)}$. For this reason, SGD can be seen as an 
\textbf{online} algorithm that processes a stream of steadily produced data 
points, one example at a time.

As expected, this gradient estimation is highly noisy, an issue that can be 
alleviated by using not one but a randomly sampled subset of learning examples. 
This subset is called 
a \textbf{mini-batch}, and its size $n$ is the \textbf{batch size}, which should 
not be confused with the above-mentioned deterministic optimizer.

Moreover, SGD has an iterative nature, and when training an ANN, one iteration is 
called \textbf{epoch}. In every epoch, the training data is divided into a 
number of mini-batches. The mini-batches are disjoint subsets of the training set, 
and the last mini-batch may have fewer examples as $n$. After the 
first mini-batch has been used to estimate the gradient, training continues with 
the next mini-batch, and so forth, until the training data has been completely 
processed. In this tutorial, our learning algorithm usually ends after a 
predefined number of epochs $\chi$.

Perhaps surprisingly, the strategy of SGD not only accelerates training but 
improve generalization as well, as concisely stated by \cite{pmlr-v48-hardt16}, 
``train 
faster, generalize better". Although SGD was first introduced by 
\cite{Robbins1951_SGD}, it is widely used up to now. For an amenable 
discussion in SGD recent trends \cite{Newton2018} may be consulted.

\subsubsection{SGD with Momentum}\label{subsubsec:sgd_with_momentum}

Despite the computational and efficacy advantages of SGD, its trajectory to the 
optimum or an appropriate stationary point can be slow. To overcome this 
limitation, the \textbf{momentum} algorithm changes the weights such as it 
proceeds incorporating the contribution of preceding gradients. This accelerates 
learning because gradients strong fluctuations can be diminished, and 
therefore, the path to the minimum is less oscillating.

One can gain intuition into the momentum algorithm by drawing a physical analogy. 
Let the learning algorithm be a heavy ball rolling down the optimization landscape 
. The ball is subject to two forces, the momentum force acting in the negative 
direction of the gradient $-\nabla_{\boldsymbol{w}} J(\boldsymbol{w})$, and a 
viscose force attenuating its movement $m \cdot v$. Therefore, the 
weights should be updated according to the net force acting on the ball.  

In particular, since momentum is mass times velocity, and considering the 
ball to have unit mass, the velocity can be interpreted as momentum. Then, 
introducing a term $\boldsymbol{v}$ and a parameter $\alpha \in [0,1)$ the 
weight update can be written

\begin{align}\label{eq:scg_with_momentum}
	\boldsymbol{v} &\leftarrow \alpha \boldsymbol{v} - \epsilon 
	\nabla_w J(\boldsymbol{w})\\\label{eq:scg_with_momentum_gradient}
	\boldsymbol{w} &\leftarrow \boldsymbol{w} + \boldsymbol{v}
\end{align}

Despite the parameter $\alpha$ being more related to \textit{viscosity}, it 
is 
widely called \textbf{momentum parameter}. Furthermore, the velocity update at 
iteration $k$ can be written as $v_k = \alpha^k v_0 - \epsilon \sum_{i = 0}^{k-1} 
\alpha^i \nabla J_w(\boldsymbol{w}_{k - 1 - i})$. Given that $\alpha < 1$, the 
velocity accumulates previous 
gradients and updates the weights using an exponential moving average, where 
recent gradients 
have more importance (lower powers of $\alpha$) than previous gradients.

SGD with momentum may be generalized to a family of optimization methods called 
`Linear First Order Methods' \cite{goh2017why}, but the presentation here shall be 
enough for our discussion.

\subsubsection{Forward Propagation and Back-propagation}\label{subsub:backprop}
In this subsection, we departure from our two main references and follow 
\cite{bishop2007} and \cite{nielsenneural}.

We are now ready to begin training the network. This requires a proper definition 
of 
an individual weight in a certain layer $l$, i.e., the weight $w^l_{jk}$ connects 
the $k^{th}$ neuron in the $(l - 1)^{th}$ layer to the $j^{th}$ neuron in the 
current layer $l$. This rather intricate notation is necessarily in order to 
compactly relate the activation of consecutive layers $\boldsymbol{a}^l$ and 
$\boldsymbol{a}^{l - 1}$:

\begin{align}\label{eq:activation_scalar}
	a^l_j &= g\left( \sum_{k} w^l_{jk} a^{l - 1}_k  + b^l_j \right)\\
	\boldsymbol{a}^l &= g(\boldsymbol{W  }^l \boldsymbol{a}^{l - 1} + 
	\boldsymbol{b}^l).\label{eq:activations_vector}
\end{align}

\begin{figure}[h]
	\centering
	\includegraphics[height=4cm]{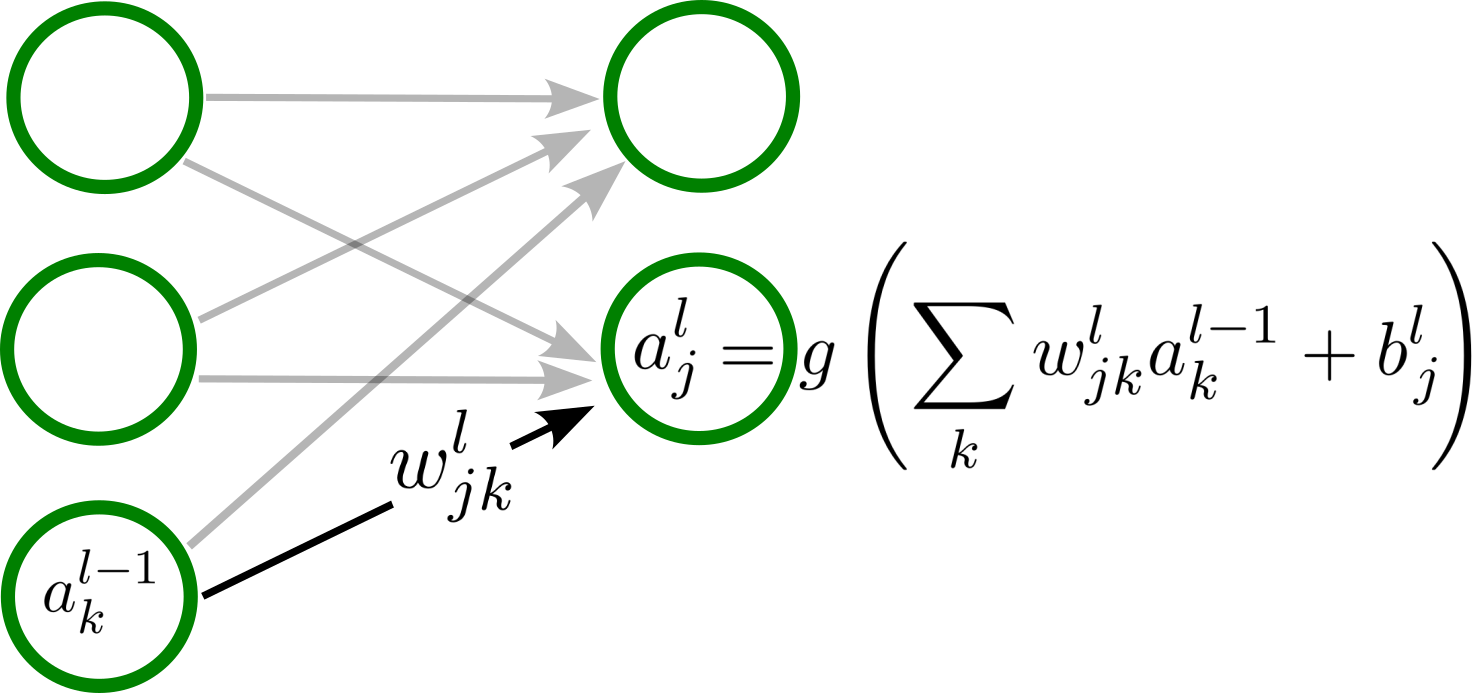}
	\caption{ANN weights notation illustrated. The figure depicts two fully 
		connected hidden 
		layers of an arbitrary 
		ANN. The layer $l$ (right) has two 
		neurons (green 
		circles) 
		and the preceding layer $l - 1$ (left) has three units. The neuron $j$ is the 
		second unit (from top to bottom) in layer $l$; the third unit $k$ in layer 
		$l-1$ has activation $a^{l - 1}_k$. Among all connections (mostly in gray), we 
		emphasize (in black) the connection from the unit $k$ in layer $l - 1$ to the 
		unit $j$ 
		in layer $l$; the weight of this connection is denoted $w^l_{jk}$. Supposing 
		the right layer is the second 
		hidden layer, then the 
		depicted weight is concretely $w^2_{23}$.
		To calculate the activation $a^l_j$ of this neuron $j$, the weights of all 
		incoming connections from all neurons $\{w^l_{j1}, w^l_{j2}, \cdots, w^l_{jk} 
		\}$ in layer $l - 1$ must be multiplied by the corresponding activation $\{a^{ 
			l - 1}_{1}, a^{l - 1}_{2}, \cdots, a^{l - 1}_{k}\}$. The bias $b^l_j$ must be 
		added to the resulting summation and the activation function $g(\cdot)$ must 
		be applied elementwise. To avoid having to transpose the weight matrix 
		representing all connections at one layer
		($(\boldsymbol{W}^l)^T$), the index order in the weights are set to be $jk$.}
	\label{fig:weights_proper_definition}
\end{figure}

Here, if we fixate the neuron $j$ with weight $w^l_{jk}$ and bias $b^l$, at layer 
$l$, then its activation $a^l_j$ 
results from the activation in the preceding layer according to Eq. 
\ref{eq:activation_scalar}. Intuitively, the activation of a neuron 
is a function of its weighted inputs, i.e., the activation of all the units 
connected to it from the previous layer, and the bias of the neuron (Fig. 
\ref{fig:weights_proper_definition}). Eq. 
\ref{eq:activations_vector} is its vectorized 
form, where $g(\cdot)$ is the activation function applied element-wise.

Furthermore, the weight sum before applying the non-linearity is called the  
\textbf{weighted input} $z^l_j$ to the neuron $j$ in layer $l$; it is defined as

\begin{align}
	z^l_j &\stackrel{\text{def}}{=} \sum_{k} w^l_{jk} a^{l - 1}_k  + b^l_j 
	\label{eq:weighted_input}\\
	\boldsymbol{z}^{(l)} &\stackrel{\text{def}}{=}  \boldsymbol{W}^{(l)} 
	\boldsymbol{a}^{(l-1)} + \boldsymbol{b}^l 
	\label{eq:weighted_input_vectorized}\\
	\boldsymbol{a}^{(l)} &= 
	g(\boldsymbol{z}^{(l)}). 
	\label{eq:activation_function_weighted_input_vectorized1}
\end{align}

Equation \ref{eq:weighted_input_vectorized} is the weighted input in vectorized 
form 
and Eq. \ref{eq:activation_function_weighted_input_vectorized1} is the activation 
written as a function of the weighted input.

Training requires \textbf{initializing the weights and 
	the bias}. More importantly, an appropriate weights initialization is crucial 
for convergence. The general idea is to break the symmetry among 
the units such as every node computes a different function.	Although there are 
several initialization strategies, 
usually the weights are drawn randomly from a Gaussian or 
uniform distribution.

However, due to the chain nature of ANN, two problems 
might arise: \textbf{vanishing gradients} or \textbf{exploiting gradients}. If 
the weights are initialized with extremely small values, e.g., from a Gaussian 
with zero mean and std. $0.01$, the hidden layer 
activations may tend to zero. Therefore, the gradient of the cost function 
may vanish. Exactly the contrary might happen if the weights are initialized 
with relative high values (it suffices with an increase of the before 
mentioned std. to $0.05$), which might lead to exceedingly large gradients.
Both vanishing and exploiting gradients hinder learning. In general, the 
problem is that large weight variance can render learning 
unstable.

\textbf{Kaiming initialization} \cite{KaimingInit2015} (also \textbf{He 
	initialization}) attempts to avoid learning 
instability by 
constraining the weights variance to be nearly constant across all layers. 
This 
strategy was first proposed by \cite{xaver_init_2010} but the derivation 
did not contemplate ReLU. Concretely, the \textit{heuristic} is to derive the 
relation by enforcing that contiguous layers have equal output variance 
$Var(\boldsymbol{a}^{(l)}) = 
Var(\boldsymbol{a}^{(l-1)})$. This set all bias to zero and the weights at 
layer $l$ to randomly 
drawn values 
from a Gaussian distribution with mean zero and std. inversely proportional to 
the 
width $\varpi^{(l - 1)}$ (number of units in layer $l - 1$, also 
\textit{fan-in}) of the 
preceding layer, i.e.,

\begin{align}\label{eq:kaimimg_init}
	\boldsymbol{b}^l &= \boldsymbol{0}\\
	w^l_{jk} &\sim  \mathcal{N}\bigg(0, \frac{2}{\varpi^{(l - 1)}}\bigg).
\end{align}

Kaiming is the default initialization implemented by Pytorch 
\cite{pytorch_2019}, the current most prominent deep 
learning framework. We also initialize both weights and bias following that 
strategy.

After the network parameters have been initialized, the \textbf{forward 
	propagation} must be performed. 
Without loss of generalization we will consider training with one example. The 
first layer receives the training example $\boldsymbol{x}^{(i)}$ as input $f^{(0)} 
= \boldsymbol{x}^{(i)}, 
\boldsymbol{a}^{(0)} = \boldsymbol{x}^{(i)}$. The activations of the hidden and 
the 
output layer may be calculated using Eq. \ref{eq:activation_scalar}.

Because the network has depth $\ell$, the activation of the last layer is the 
network's output $\boldsymbol{a}^{\ell} = f^{\ell}(\boldsymbol{x}^{(i)})$, which 
can be 
regarded, without loss of generalization, as a vector $\hat{\boldsymbol{y}}_i$ for 
some 
example $\boldsymbol{x}^{(i)}$. 
This allows to execute completely a forward pass. Before calculating the loss, let 
us first discuss the weight update.

Ultimately, learning means that weights must be updated with 
Eqs.\ref{eq:scg_with_momentum}-\ref{eq:scg_with_momentum_gradient}. 
\textbf{Back-propagation} (backprop) is the method that enables that update. Note 
that the term $\nabla_w J(\boldsymbol{w})$ in Eq. \ref{eq:scg_with_momentum} is the gradient of the cost 
function 
with respect to every weight $w^l_{jk}$ in the network. As we will now see, 
calculating the gradient at the output layer is straightforward. The problem is 
that calculating the gradient of the cost function $J$ at hidden layers is not 
obvious. The goal of backprop is to calculate the gradient of the cost function 
w.r.t. every weight $w^l_{jk}$ in the network.

To achieve that goal, two important assumptions need to be made. First, the cost 
function $J$ can be decomposed into separate cost functions for individual 
training examples $\boldsymbol{x}^{(i)}$. This is indeed the case for MSE in Eq. 
\ref{eq:mse}. 
This assumption is 
needed because it reduces the gradient problem to determine the gradient for one 
training example $\nabla_{w} J_i(\boldsymbol{w})$, for some general training 
example 
$\boldsymbol{x}^{(i)}$. Then the 
gradient can be totalized by averaging over all 
training examples

\begin{equation}\label{eq:gradient_averaging}
	\nabla_{w} J(\boldsymbol{w}) = \frac{1}{m} \sum_{i=1}^{m} 
	\nabla_{w} J_i(\boldsymbol{w}).
\end{equation}

Second, the cost function must be written as a function of the outputs; otherwise, the computation of the derivative at the output layer would not be possible \cite{nielsenneural}.

With those assumptions, the goal of backprop can be restated as calculating the 
gradient of the cost function w.r.t. every weight $w^l_{jk}$ in the network for an 
individual training example $\boldsymbol{x}^{(i)}$.

The general idea is to determine four equations. First, calculate the error at 
the output layer. Second, calculate the total error of an arbitrary hidden layer 
in terms of the next layer, which allows to \textbf{back-propagate} the error from 
the output to 
the input layer. Third, calculate the derivative of the cost function w.r.t the 
bias. Finally, the first two equations are combined to calculate the 
gradient for every weight in the network.

All four equations may be derived applying the \textbf{chain rule of 
	calculus} 
for a 
variable $z$ transitively depending on variable $x$, via the intermediate 
variable 
$y$ as $z = f(y)$ and $y = g(x)$. The chain rule of calculus relates the 
derivative of $z$ with respect to $x$ as in $\diff{z}{x} = \diff{z}{y} 
\cdot \diff{y}{x}$. Since the cost is a function of all weights, partial 
derivatives 
must be used  $\frac{\partial z}{\partial x} = \frac{\partial z}{\partial y} 
\cdot \frac{\partial y}{\partial x}$.

A quantity that we will need is the \textbf{error at an individual unit} 
(sometimes called \textbf{neuron error} or \textbf{sensitivity})
$j$ in layer $l$, written as $\delta^i_j$. The intuition is that the error at the 
output layer ( 
quantified by the cost function $J$) varies if the weighted input of one 
arbitrary node in an arbitrary hidden layer is perturbed by certain amount 
$\Delta$. This perturbation disseminates through all connected nodes until it 
reaches the output nodes, where the error is calculated. Therefore, the partial 
derivative of the error with respect to the weighted input of this unit may be 
interpreted as a measure of the error attributable to this neuron. Indeed, 
$\diffp[]{J}{{z^l_{j}}}$ expresses how much the error varies, when the 
weighted 
input is perturbed by a small amount $\Delta z^l_{j}$. Therefore, the neuron 
error may be defined as

\begin{equation}\label{eq:neuron_error}
	\delta^l_{j} \stackrel{\text{def}}{=} \diffp[]{J}{{z^l_{j}}}
\end{equation}

Note that at the output layer, $J$ is a function of all activations $a^\ell_i$. 
Its 
rate 
of change  
with respect to a weighted input $\frac{\partial J}{\partial z^\ell_j}$ depends on 
all contributions 
that the weighted input influences. Formalizing that intuition and applying the 
generalized chain rule of 
calculus, we have

\begin{equation}\label{eq:error_output_layer_sum}
	\delta^\ell_{j} = \sum_k \diffp[]{J}{{a^\ell_{k}}} 
	\diffp[]{{a^\ell_{k}}}{{z^\ell_{j}}}
\end{equation}

where the summation is over all output units $n$. In the case of MSE, the weighted 
input 
$z^\ell_j$ influences only $a^\ell_j$. Therefore, the other partial derivatives 
($i 
\neq j$) 
vanish and the equality becomes 

\begin{align}\label{eq:error_output_layer}
	\delta^\ell_{j} &= \diffp[]{J}{{a^\ell_j}} 
	\diffp[]{{a^\ell_j}}{{z^\ell_j}}\\
	&=\diffp[]{J}{{a^\ell_j}} \cdot g^\prime(z^\ell_j)\\
	\boldsymbol{\delta}^\ell &= \nabla_a {J} \odot g^\prime(\boldsymbol{z}^\ell), 
	\label{eq:error_output_layer_matrix_form}
\end{align}

being $g^\prime(z^\ell_j)$ the derivative of the activation function $g(\cdot)$ 
evaluated at node $j$. 
Eq. 
\ref{eq:error_output_layer_matrix_form} is the matrix form, where $\nabla_a {J}$ 
is the 
vector of partial derivatives of the cost function w.r.t. the activations, and  
$\odot$ is the 
Hadamard product (elementwise multiplication). 

Subsequently, the error relating two contiguous layer $l$ and $l + 1$ must be 
defined. Similar to the derivation of the error at the ouput layer, the chain rule 
must be applied. However, now the cost function must be considered to be dependent 
on the 
weighted input at layer $z^l_j$ throughout intermediate weighted inputs at the 
next layer $z^{l + 1}_j$ for all the neurons to which the unit $j$ connects. That 
is

\begin{align}
	\delta^l_{j} &= \diffp[]{J}{{z^l_j}}\\ 
	\delta^l_{j} &= \sum_{k} \diffp[]{J}{{z^{l + 1}_k}} 
	\diffp[]{{z^{l + 1}_k}}{{z^l_j}} \label{eq:error_hidden_layer_sum_1}\\
	&= \sum_{k} \diffp[]{{z^{l + 1}_k}}{{z^l_j}} \delta^{l + 1}_k. 
	\label{eq:error_hidden_layer_sum_2}
\end{align}

Substituting the first therm on the right in Eq. 
\ref{eq:error_hidden_layer_sum_1} 
by its definition $\delta^{l + 1}_k$ and interchanging with the second term 
results in Eq. \ref{eq:error_hidden_layer_sum_2}. 

Taking as a reference the unit $k$ in the $l + 1$ layer, the summation runs over  
all units $j$ on the layer $l$ that link to the unit $k$. Thefore,

\begin{align}
	z^{l + 1}_k &= \sum_{j} w^{l + 1}_{kj} a^{l}_j  + b^{l + 1} 
	\label{eq:weighted_input_layer_l_plus_one}\\
	z^{l + 1}_k &= \sum_{j} w^{l + 1}_{kj} g(z^l_j)  + b^{l + 1} 
	\label{eq:weighted_input_layer_l_plus_one_1}\\
	\diffp[]{{z^{l + 1}_k}}{{z^l_j}} &= w^{l + 1}_{kj} 
	g^\prime(z^l_j),\label{eq:weighted_input_layer_l_plus_one_2}
\end{align}

where we substitute in Eq. \ref{eq:weighted_input_layer_l_plus_one} the 
activation 
by its corresponding nonlinearity (Eq. \ref{eq:weighted_input_layer_l_plus_one_1}) 
and further derive to obtain Eq. \ref{eq:weighted_input_layer_l_plus_one_2}. This 
equation may be inserted in Eq. \ref{eq:error_hidden_layer_sum_2} resulting in

\begin{align}
	\delta^l_{j} &= \sum_k w^{l + 1}_{kj} \delta^{l + 1}_k
	g^\prime(z^l_j) \label{eq:final_layer_one_layer_minus_one}\\
	\boldsymbol{\delta}^l &= \big((\boldsymbol{W}^{l + 1})^T 
	\boldsymbol{\delta}^{l + 1}\big) 
	\odot 
	g^\prime(\boldsymbol{z}^l), \label{eq:final_layer_one_layer_minus_one_matrix}
\end{align}

where $(\boldsymbol{W}^{l + 1})^T$ is the transpose of the weight matrix 
$\boldsymbol{W}^{l + 1}$ and Eq. 
\ref{eq:final_layer_one_layer_minus_one_matrix} is the 
matrix 
form. This 
equation indicates that, having obtained the weight matrix at an arbitrary current 
layer by forward propagation, the error at the layer immediately before can be 
calculated by 1) multiplying the transpose of that weight matrix by 
he 
error at 
the current layer and 2) taking the Hadamard product of the resulting 
vector.

Having established the error at the output layer and the backpropagation rule, it 
remains to determine the relation of the partial derivative of the cost function 
w.r.t. any individual weight at an arbitrary layer, in terms of the error at that 
layer $\delta^l$. That is

\begin{align}
	\diffp[]{J}{{w^l_{jk}}} &= \diffp[]{J}{{z^l_j}} \diffp[]{z^l_j}{{w^l_{jk}}}\\
	&= \diffp[]{z^l_j}{{w^l_{jk}}} \delta^l_j  
	\label{eq:partial_with_respect_to_weight_1}\\
	&= \diffp[]{\sum_{k} w^l_{jk} a^{l - 1}_k  + b^l}{{w^l_{jk}}} \delta^l_j 
	\label{eq:partial_with_respect_to_weight_2}\\
	\diffp[]{J}{{w^l_{jk}}} &= a^{l - 1}_k \delta^l_j. 
	\label{eq:partial_with_respect_to_weight_3}
\end{align}

Here, again the terms in the right have been interchanged, and the first term in 
the right has been substituted by the error definition in Eq. 
\ref{eq:partial_with_respect_to_weight_1}. In Eq. 
\ref{eq:partial_with_respect_to_weight_2} the weighted input $z^l_j$ has been 
expanded, and the derivation realized, resulting in Eq. 
\ref{eq:partial_with_respect_to_weight_3}. Note that $a^{l - 1}_k$ is the 
activation of the neuron in the layer immediately before that links the current 
neuron, a quantity that has been calculated by forward propagation. The second 
term is the current neuron error (Eq. \ref{eq:final_layer_one_layer_minus_one}) 
that has been recursively 
backpropagated from the output layer.

Finally, the gradient of the cost function w.r.t. the bias is

\begin{align}
	\diffp[]{J}{{b^l_j}} &= \diffp[]{J}{{z^l_j}} \diffp[]{z^l_j}{{b^l_j}} 
	\label{eq:gradient_cost_wrt_bias_chain}\\
	\diffp[]{J}{{b^l_j}} &= \delta^l_j, \label{eq:gradient_cost_wrt_bias}
\end{align}

where the chain rule has been applied (Eq.  
\ref{eq:gradient_cost_wrt_bias_chain}) and the error at the layer has been 
replaced 
by its definition, resulting in Eq. \ref{eq:gradient_cost_wrt_bias}. In words, the 
contribution of the bias to the gradient equals the neuron error.

The four main equations above (Eqs. 
\ref{eq:error_output_layer_matrix_form}, 
\ref{eq:final_layer_one_layer_minus_one_matrix}, 
\ref{eq:partial_with_respect_to_weight_3}, and \ref{eq:gradient_cost_wrt_bias}) 
enable calculating the gradient of the cost function w.r.t. every weight and bias. 
As 
already mentioned, in the context of deep learning, this is exactly the goal of 
backprop. The derivatives may then be used to train the network by updating the 
weights. Algorithm \ref{algo:ann_training} shows the complete training 
procedure.
For a general derivation of backpropagation using \textbf{computational graphs} 
\cite{Goodfellow.2016} may be consulted.

%One fundamental characteristic of backprop is its efficiency. The naive approach 
%to obtain the individual partial derivatives of the gradient vector would 
%approximate those gradients
\begin{figure}[H]
	\vspace{-1cm}
	\begin{algorithm}[H]
		\DontPrintSemicolon
		
		\KwInput{A set of ordered training examples $\mathbb{X}$ with targets 
			$\mathbb{Y}$, an ANN of depth $\ell$.}
		\KwOutput{A trained ANN (model) $\hat{f}(\cdot)$.}
		
		\textbf{Initialization:}
		Set parameters according to Kaiming initialization (Eq. 
		\ref{eq:kaimimg_init}).\;
		Set the hyperparameters (Table \ref{tbl:hyperparameters}).\;% to their 
		%concrete 
		%values and designate an activation function $g(\cdot)$.\;
		
		\textbf{Iterative minimization of the cost function with gradient descent:}\;
		\ForEach{epoch $e \in {1, \cdots, \chi}$}{	
			Randomly generate mini-batches $\mathbb{B}^{(i)}$ of size $n$.\; %The 
			%mini-batches are disjoint subsets of the training examples. The last 
			%mini-batch may contain fewer elements as $n$.\;
			\textbf{Specifically use stochastic gradient descent:}\;
			\ForEach{Mini-batch $\mathbb{B}^{(i)}$}{

				\ForEach{Example $\boldsymbol{x}^{(i)}$ in mini-batch and its 
					corresponding target $\boldsymbol{y}_i$}{
					
					\textbf{Input the training example:}\;
					The activations of the input layer are considered to be the example 
					data.\;
					$\boldsymbol{a}^{(0)} \leftarrow \boldsymbol{x}^{(i)}$
					
					\textbf{Perform the forward propagation pass:}\;
					\ForEach{Layer $l \in \{1, \cdots, \ell\}$}{
						Compute weighted input and activations (Eq. 
						\ref{eq:weighted_input_vectorized}, 
						\ref{eq:activation_function_weighted_input_vectorized1}).\;
						$\boldsymbol{z}^{(l)} =	\boldsymbol{W}^{(l)} 
						\boldsymbol{a}^{(l-1)} + \boldsymbol{b}^l$\;
						$\boldsymbol{a}^{(l)} \leftarrow g(\boldsymbol{z}^{(l)})$\; 
					}
					\textbf{Calculate the error vector at the output layer:} 
					(Eq.			
					\ref{eq:error_output_layer_matrix_form})\;
					$\delta^\ell = \nabla_a {J} \odot g^\prime(\boldsymbol{z}^\ell)$\;
					\textbf{Backpropagation:}\;
					\ForEach{Layer $l \in \{\ell, \cdots, 1\}$}{
						Backpropagate the error (Eq. 
						\ref{eq:final_layer_one_layer_minus_one_matrix}).\;
						$\delta^l = \big((\boldsymbol{W}^{l + 1})^T \delta^{l + 
							1}\big) \odot 
						g^\prime(z^l)$\;
						Calculate local gradients (Eq. 
						\ref{eq:partial_with_respect_to_weight_3})\;
						$\diffp[]{J}{{w^l_{jk}}} = a^{l - 1}_k \delta^l_j$\;
						This is the gradient corresponding to one training exaple.\;
						$\nabla_{w} J_i(\boldsymbol{w}) = \{\diffp[]{J}{{w^1_{jk}}}, 
						\cdots, 
						\diffp[]{J}{{w^\ell_{jk}}}\}$			
						
					}
					\textbf{Recover all gradients:}\;
					Average over all examples (Eq. \ref{eq:gradient_averaging}).\;
					$\nabla_{w} J(\boldsymbol{w}) = \frac{1}{n} \sum_{i=1}^{n} 
					\nabla_{w} J_i(\boldsymbol{w})$.\;
					\textbf{Apply the weight update:}\;
					Update the weight (Eq. \ref{eq:gradient_descent}).\;
					$\boldsymbol{w} \leftarrow \boldsymbol{w} - \epsilon \nabla_w 
					J(\boldsymbol{w})$
			}}	
		}

		\caption{Example ANN training algorithm. Here, we implement training by 
			iterating over examples in a mini-batch. In practice, backprop is implemented 
			to computed simultaneously the gradients of the entire mini-batch using 
			matrices. 
		}\label{algo:ann_training}
	\end{algorithm}
\end{figure}

\pagebreak

\subsubsection{Hyperparameters}

In most discussions through this tutorial, we have mentioned several quantities, but we did not indicate how they can be set. These quantities are called \textbf{hyperparameters}.

Hyperparameters are usually set empirically or using some kind of 
\textit{intuition}. Finding suitable values of these hyperparameters is termed Hyperparameter \textbf{tuning}. Arguably, the easiest way to establish their values is to set their \textit{default values} or 
\textit{common values}. Table \ref{tbl:hyperparameters} summarizes them.

Automated hyperparameters tuning algorithms (also \textbf{AutoML} - Automated Machine Learning) include \textbf{grid search}, \textbf{random search}, and \textbf{model-based hyperparameter optimization}. A comprehensive discussion of these methods may be found in \cite{hutter2019automated}.

{\def\arraystretch{1.5}\tabcolsep=8pt
	\begin{table}[h!]
		\centering
		\begin{tabular}{|c|c|c|c|}
			\toprule
			Hyperparameter & Interpretation & Section & Default or common values \\
			\hline
			$\epsilon$ & Learning rate & \ref{subsubsec:gradient_descent} & 
			$\mathnormal{0.01}$, $\mathnormal{0.1}$ \\
			%\hline
			$\alpha$ & Regularization coefficient & \ref{subsec:regularization} & 
			$\mathnormal{0.2}$ \\
			%\hline
			$\chi$ & Number of epochs & \ref{subsubsec:sgd} & $\mathnormal{5}$, 
			$\mathnormal{10}$, $\mathnormal{100}$ \\
			%\hline
			$n$ & Batch size (SGD) & \ref{subsubsec:sgd} & $\mathnormal{100}$ \\
			%\hline
			$\upsilon$ & Momentum &  \ref{subsubsec:sgd_with_momentum} & 
			$\mathnormal{0.9}$ \\
			\bottomrule
		\end{tabular}
		\caption{Hyperparameters. The table summarizes the hyperparameters that we use in 
			this 
			tutorial (first column). The second column indicates the corresponding 
			interpretation. Furthermore, we specify in which subsection they are described 
			(third column) and their \textit{typical} values, in the fourth column.}
		\label{tbl:hyperparameters}
	\end{table}
}

\section{Convolutional Neural Networks}\label{sec:cnn}
In Sec. \ref{subsubsec:deep_forward_networks}, we mentioned that the architecture 
of a general ANN may involve 
different layer types. Specifically, we described a dense 
layer, whose mathematical form is basically matrix multiplication. Unfortunately, 
an architecture composed alone of dense layers is computational 
extremely inefficient. For example, processing $200 \times 
200$ pixels monochrome images, like the ones we input to the ANN, requires $40000$ 
connections (one connection to every pixel) only for one neuron in the first 
hidden layer.

\textbf{Convolutional Neural Networks} (\textbf{CNN}, also \textbf{ConvNets}) 
reduce strikingly the computational burden of having to store and update millions 
of weights in the network. Two characteristics convert CNNs in an 
effective learning agent, namely 
\textbf{sparse interactions} (also \textbf{sparse connectivity} or \textbf{sparse 
	weights}) and \textbf{parameter sharing}, which we will discuss 
beneath.

Curiously, the 
invention of CNNs was not motivated by efficiency considerations. Instead, 
ConvNets, like ANNs and Batchnorm (which we discuss further in \ref{subsec:batch_norm} ), are inspired by the functioning of the brain: 
``the network has a
structure similar to the hierarchy model of the visual
nervous system. ($\cdots$) The structure of this network has been
suggested by that of the visual nervous system of the
vertebrate. " \cite{Fukushima1980NeocognitronAS}. In particular, the vertebrate brain exhibits a hierarchical structure where cells higher in the hierarchy (complex cells) respond to more complex patterns, while cells at lower stages (simple cells) respond to simpler patterns. Complex cells are also more robust to shifts in the input, a property now referred to as \textbf{equivariant representation}.

\begin{figure}[H]
	\centering
	\includegraphics[width=14.6cm]{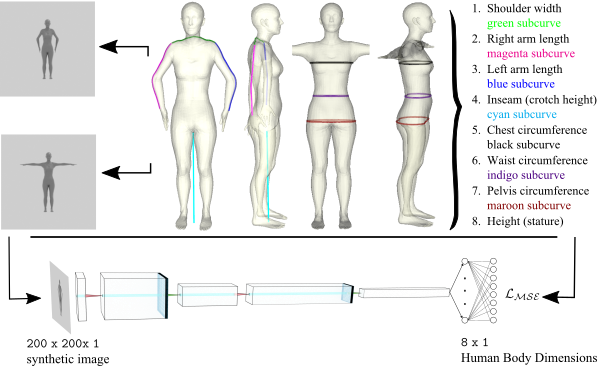}
	\caption{An example of a CNN architecture performing regression. Here, we show an adapted version of our Neural Anthropometer from \cite{gon_mayer_na}, a CNN that is able to regress eight human body dimensions like height or waist circumference from synthetic images of persons in two poses (top left). The input to the network are grayscale $200 \times 200$ pixels synthetic images of 3D real human meshes (top middle). The supervision signal is a vector of eight human body dimensions (top right) and the loss $\mathcal{L}_{\mathcal{MSE}}$ (bottom right) is the Mean Square Error MSE (Eq. \ref{eq:mse}) between the actual and the estimated measurements. Hyperparameters for learning with this CNN are set as described in Table \ref{tbl:hyperparameters}. We discuss the details of the layer types in Subsections \ref{subsec:conv_layer}-\ref{subsec:relu}. At the bottom, the five gray rectangular cuboids represent the input and the feature maps. Red and green horizontal pyramids represent convolution (Subsec. \ref{subsec:conv_layer}) and max pooling (Subsec. \ref{subsec:pooling}) layers respectively. Black grids are Rectified linear unit (ReLU) layers (Subsec. \ref{subsec:relu}), and connected circles are fully connected layers (Subsec. \ref{subsubsec:deep_forward_networks}). In this case, 1) the first convolutional layer applies a convolution with a $5$-pixels square kernel to produce a feature map of size $196 \times 196 \times 8$. The tensor is then passed through a ReLU and batch normalization (Subsec. \ref{subsec:batch_norm}, not shown here) is applied. Next, 2) max pooling  with stride 2 is used producing a tensor of size $98 \times 98 \times 8$. Then, 3) the tensor is send to a second convolutional layer with a $5$-pixels square kernel and $16$ output channels, resulting in a tensor of size $94 \times 94 \times 16$. Following, 4) pooling is applied as before producing a tensor of size $47 \times 47 \times 16$ and 5) the output is flatten to a tensor of size $35344$. This tensor is passed to a fully connected layer and again through a ReLU (not shown). The output layer 6) regresses the eight human body dimensions in meters. Using the simple layer terminology, the depth of this network is $10$: convolution + ReLU + Batchnorm + max pooling + convolution + ReLU + max pooling + fully connected + ReLU + fully connected (output layer). Other common counting schema includes only the layers with learnable parameters, in which case this network would have depth $5$.}
	\label{fig:neural_anthro}
\end{figure}

ConvNets have their own distinctive terminology. In particular, at least one of 
its layers is a \textbf{convolutional layer} (Subsec. \ref{subsec:conv_layer}). 
To describe the components of a CNN, we adopt the so-called \textbf{simple 
	layer 
	terminology} (in opposed to complex layer terminology). For example, we 
consider that the convolutional layer does not apply a non-linearity. 
Therefore, the layer's output is the addition of the convolution 
\ref{eq:cross_correlation} and a bias $b$, termed \textbf{pre-activation}. This 
allows us to have a finer control 
of the description that will be useful when explaining the concrete CNN 
architecture. Note that in this terminology 
not every layer has weights, e.g., Subsec. \ref{subsec:pooling}. Fig. \ref{fig:neural_anthro} shows an example of CNN that is able to estimate eight human body dimensions from grayscale synthetic images.

\begin{figure}[H]
	\centering
	\includegraphics[width=14.6cm]{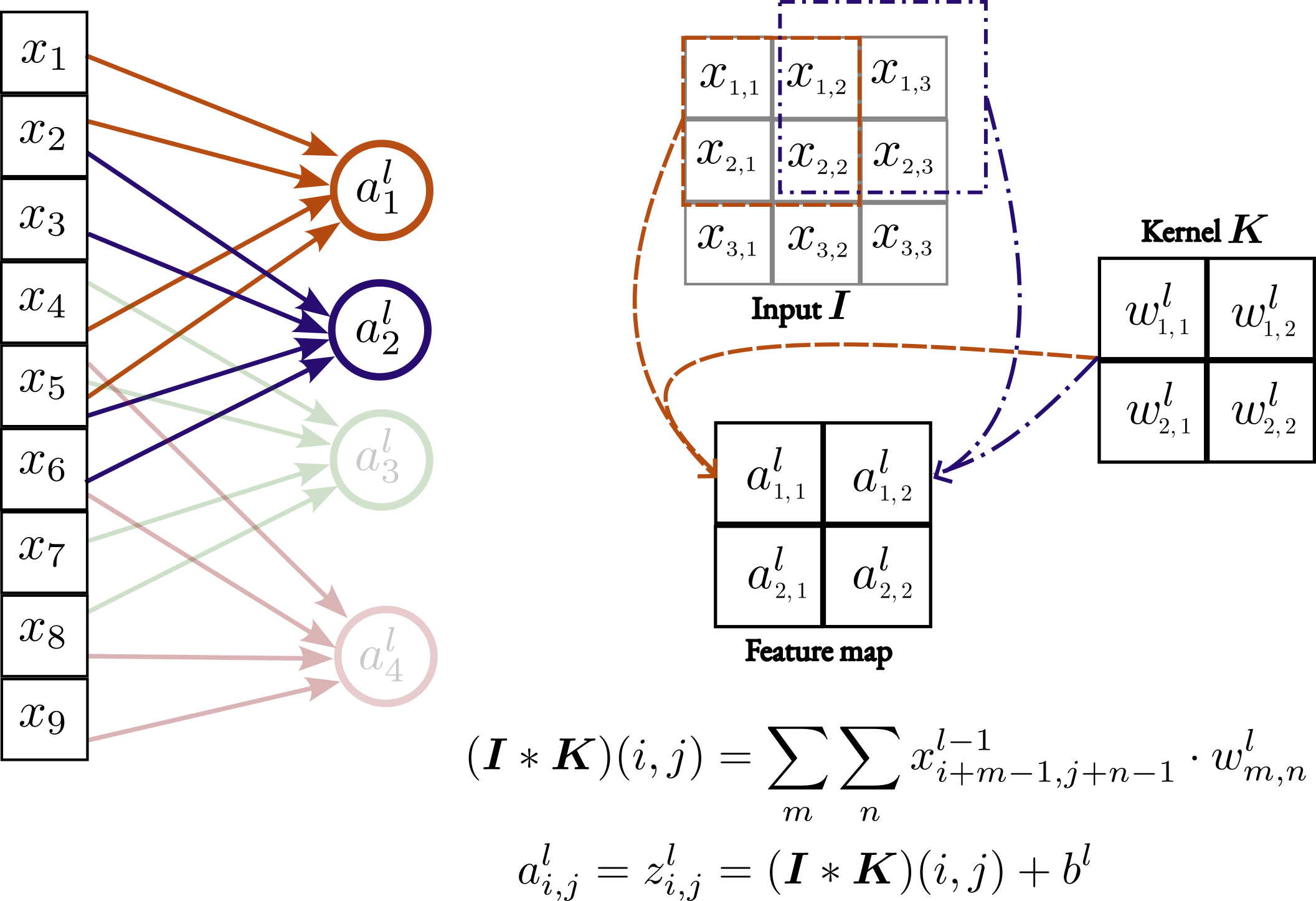}
	\caption{Possible input and first hidden layer of a hypothetical CNN. Left, 
		A general ANN that receives the input $\{x_1, \cdots, x_9\}$. The first 
		hidden layer $l$ 
		contains 
		four neurons with activation $\{a_1, \cdots, a_4\}$ (circles in orange, 
		magenta, green and red color, from top to bottom). We emphasize the 
		first 
		two 
		neurons and its connections to the input layer. Note that the neurons 
		are 
		not 
		fully connected to the input layer, e.g., the unit $a^l_1$ is connected 
		to 
		inputs $\{x_1, x_2, x_4, x_5\}$, which is its receptive field, but not 
		to 
		input $x_3$. Moreover, all four 
		neurons have equal weights $\{w_1, \cdots, w_4\}$, indicated by the 
		four 
		arrows 
		arriving to the units. Right, the same network seen as a CNN. The 
		input, 
		the 
		weights and the neurons have a grid structure. The connections of the 
		neurons 
		to the input induce a $2D$-convolution $(I * K)(i,j)$ of the input 
		$I$ 
		($3 
		\times 3$ 
		grid) with 
		the 
		kernel $K$ ($2 \times 2$ grid). The convolution may be visualized as 
		sliding 
		the kernel over the input grid to produced a feature map, where every 
		element 
		of the feature map is the Frobenius product $\langle K, A \rangle_F$ of the matrix 
		defined by the kernel $K$ and the overlapped input elements $A$, plus 
		one shared bias. We 
		highlight 
		the 
		computation of two elements of the feature map (the other two are not 
		depicted) by the corresponding broken
		(\textcolor{OrangeCNNNeuron}{$-\;-\;-$}) and broken with point 
		(\textcolor{MagentaCNNNeuron}{$-\cdot-\cdot-$}) arrows of the 
		highlighted orange and 
		magenta neurons, respectively. For example, the 
		unit 
		$a^l_{1,1}$ has 
		pre-activation $z^l_{1,1} = x^l_{1,1} \cdot w^{l - 1}_{1,1} + \cdots + 
		x^l_{2,2} \cdot w^{l - 1}_{2,2} + b^l$. Note that there is only one 
		bias $b^l$ per feature map. We adopt the simple layer 
		terminology, where $a^l_{i,j}$ is a linear unit that outputs a 
		pre-activation $z^l_{i,j}$. Therefore, $a^l_{i,j} = z^l_{i,j}$. 
		Supposing that 
		the 
		input $I$ 
		is an image, it may be observed that the sparse connections and the 
		shared weights confer the network the capability of establishing 
		spatial 
		relations among neighbor pixels. For an input of size $n \times n$ 
		convolved with a $f \times f$ kernel, the size of the feature map is $n 
		- 
		f + 1 \times n - f + 1$.}
	\label{fig:cnn}
\end{figure}

Both the input to the network as the input to a specific convolutional layer 
may 
be named \textbf{input map}, while a specific set of shared weights is named as 
\textbf{kernel} (also \textbf{filter}) after the second argument of the 
convolution operation \ref{eq:cross_correlation}. Because the kernel can be seen 
as a 
\textbf{feature detector}, 
the output (activations) of the convolutional layer is called a \textbf{feature 
	map} (also \textbf{output map}). Moreover, from the perspective of one fixed 
unit $a$, the input elements affecting the calculation of its activation are 
called the \textbf{receptive field} of $a$. Fig. \ref{fig:cnn} depicts a 
possible input 
and first hidden layer of a hypothetical $2D$-CNN.

Returning to the efficiency of ConvNets, the sparse connectivity implies, 
depending on the filter size, a dramatic reduction of the network weights. Kernels 
are designed such as their size is significantly smaller than the output. In 
the 
example of the above mentioned $200 \times 200$ pixels image as input, a kernel size of \textit{only} 
$5 \times 5$ may be appropriate. Here, the reduction of the needed computational 
resources falls from $40000$ to $25$ connections per neuron in the first hidden layer. Surely, 
detecting several features requires 
a larger number of kernels. For example, popular modern CNNs like AlexNet\cite{krizhevsky2012imagenet} and VGG16 \cite{simonyan2014very}, depending on the network depth and design, employ dozens to hundreds of kernels per layer.

At the same time, the parameters (weights) in a convolutional layer are shared by 
several 
neurons, which results in a network having \textbf{tied weights}. Fully connected 
layers can 
not exploit the grid structure of the data 
because every connection is equally rated by the network. For example, if the 
input 
is an image, nearby pixels may represent a potential important feature like an 
edge or texture. However, a dense 
layer does not connect these nearby pixels in any special manner. Constraining the 
connection scheme to groups of weight sharing neurons, effectively allows to 
establish relationships between spatially related pixels.

\subsection{Convolutional Layer}\label{subsec:conv_layer}

\begin{figure}[H]
	\centering
	\includegraphics[width=14.0cm]{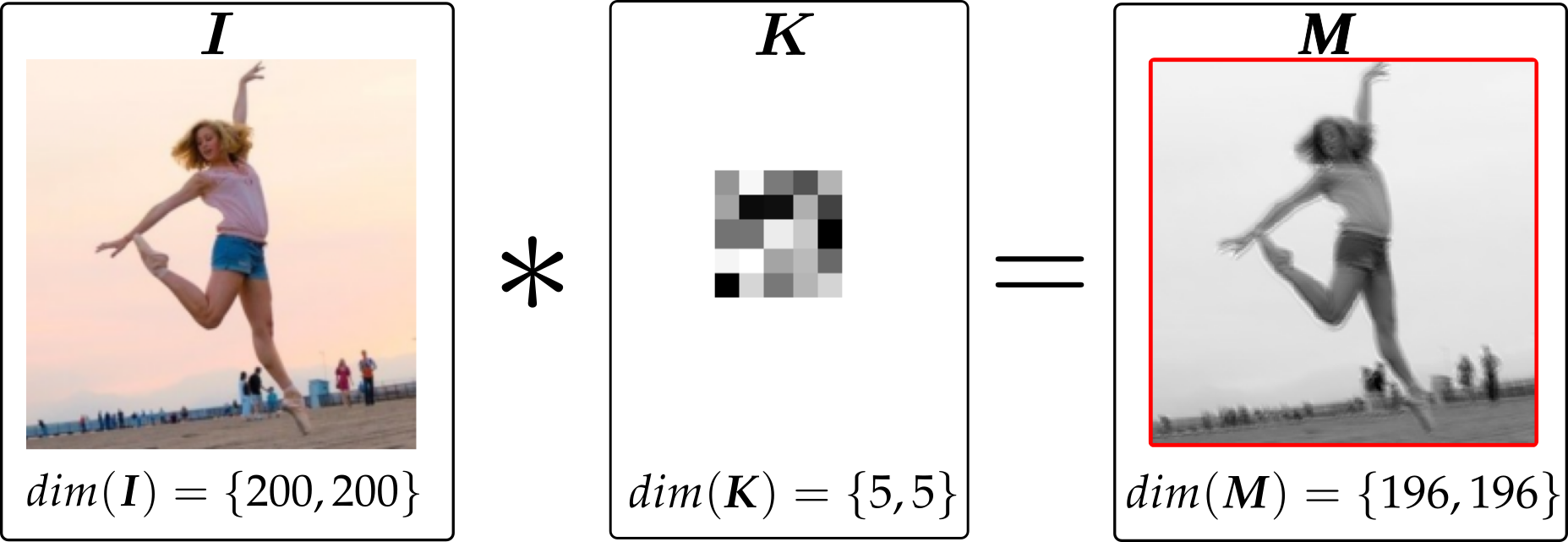}
	\caption{From left to right, the rectangles represent the concepts of the input $\boldsymbol{I}$ being convolved with a filter $\boldsymbol{F}$ and the resulting feature map $\boldsymbol{M}$ of a CNN ($\boldsymbol{I} \ast \boldsymbol{F} = \boldsymbol{M}$). In order to exemplify these concepts, we modified our Neural Anthropometer network \cite{gon_mayer_na}, that originally accepts as input a grayscale image, to accept a RGB image of $dim(\boldsymbol{I}) = \{200,200\}$ (left). On the bottom of the rectangles we indicate the dimensions of the corresponding tensor, as defined by the function $dim(\boldsymbol{X})$ (Eq. \ref{eq:size_of_feature_map} - \ref{eq:size_of_feature_map_1}). We display the first filter with $dim(\boldsymbol{F}) = \{5,5\}$ of the first convolutional layer in the trained network. Note that, in order to enhance the visibility, we significantly scale the filter in the middle. On the right, the red border indicates the reduced dimensionality of $\{4,4\}$ of the feature map with $dim(\boldsymbol{M}) = \{196,196\}$ w.r.t. the input.}
	\label{fig:input_filter_feature_map_of_cnn}
\end{figure}

Convolutional networks are named as such because performing forward propagation in 
one of their layers $l$, with sparse and tight 
weights, implies calculating a convolution of the input $\boldsymbol{I} \in 
\mathbb{R}^{m \times 
	n}$ to the layer $l$, with at least one kernel $\boldsymbol{K} \in 
\mathbb{R}^{f 
	\times f}$ of 
a determined odd squared size $f$. The result of that convolution is a feature 
map 
$\boldsymbol{M}$ (Fig. \ref{fig:input_filter_feature_map_of_cnn}).

The reason to use an odd-sized kernels (such as $3 \times 3$  or $5 \times 5$) is because they provide a well-defined center element, allowing the filter to align symmetrically with the input at each position. Additionally, they make it possible to apply equal padding on all sides, preserving the spatial dimensions of the input in the resulting feature maps.

Assume a general function $dim(\boldsymbol{X})$ that maps a matrix  
$\boldsymbol{X}$ to its dimensions. Then the dimensions of the input, kernel, 
and 
feature map are     

\begin{align}\label{eq:size_of_feature_map}
	dim(\boldsymbol{I}) &\stackrel{\text{def}}{=} \{m, n\}\\
	dim(\boldsymbol{K}) &\stackrel{\text{def}}{=} \{f, f\}\\
	dim(\boldsymbol{M}) &= \{m - f + 1, n - f + 
	1\}.\label{eq:size_of_feature_map_1}
\end{align}

In deep learning, the term convolution usually refers to the mathematical 
operation of \textbf{discrete cross-correlation}, which is the operation 
implemented by 
most deep learning frameworks. Assuming an input map $\boldsymbol{I}$ and a 
kernel $\boldsymbol{K}$, 
one-based 
index cross-correlation takes the form

\begin{align}
	{\big((\boldsymbol{I} * \boldsymbol{K})(i,j)\big)}^l &= \sum_{m} \sum_{n} 
	x^{l - 1}_{i + m - 1, j + n 
		-1} \cdot 
	w^l_{m,n}\label{eq:cross_correlation}\\
	z^l_{i,j} &= {\big((\boldsymbol{I} * \boldsymbol{K})\big)}^l + b^l_{i,j}\\
	a^l_{i,j} &= z^l_{i,j},
\end{align}

where $b^l_{i,j}$ is the bias of the pre-activation $z^l_{i,j}$ and 
$a^l_{i,j} = z^l_{i,j}$ the $i,j$ element of a $2D$ feature map.

Furthermore, the convolution operation can be generalized in a number of ways. 
First, the manner in which the kernel is slid over the input map. The kernel must 
not necessarily be moved to the immediately next spatial element to compute the 
Frobenious product. Instead, the kernel may be slid a number $s$ of spatial 
locations. This is called the \textbf{stride} $s$, and the corresponding operation 
may be called a \textbf{strided convolution}. The baseline convolution has $s 
= 1$.

Second, the kernel may be shifted beyond the 
elements 
at the border of the input map. Sliding the kernel, such as it entirely lies 
within 
the input map, has the advantage that all the elements of the feature map are a 
function of equal numbers of elements in the input. However, it also has the 
disadvantage of progressively reducing the output (Eq. 
\ref{eq:size_of_feature_map_1}) until the extreme 
stage where a $1 \times 1$ output can not be meaningfully further processed. Alternatively, the spatial dimensions may be preserved by 
\textbf{zero-padding} the input with $p$ elements. When $p = 0$ (no padding), the 
operation is called \textbf{valid convolution}, which is one of the most common. In 
contrast, constant 
spatial dimensionality across input and output may be achieved by performing a 
\textbf{same convolution} with the corresponding $p > 0$ padding.

Finally, the convolution operation may be generalized to operate on 
multidimensional inputs, 
e.g., RGB images, producing multidimensional outputs. This generalized 
convolution is termed \textbf{convolutions over volumes} because input, 
kernels, and output may be 
considered as having a width, height, and depth (also \textbf{channel} to avoid 
confusion 
with the network's depth). This, besides the processing of 
batches (Subsec. 
\ref{subsec:batch_norm}), results in CNNs usually operating in tensors of size 
$b 
\times w \times h \times c$, i.e., batch size $b$, two spatial dimensions width $w$
and 
height $h$, and $c$ number of channels, respectively. In this tutorial, we 
focus on inputs (images) with a maximum of three channels. Without loss of 
generality, we will consider one instance batches $b = 1$, therefore, we will 
omit the batch dimension. The input and the 
filters are required to have an equal number of channels $c$. The dimensions of 
the 
tensor of feature maps $\boldsymbol{\mathsfit{M}}$ are determined by the 
dimensions of the input, the size, and 
the number 
of kernels.

Regarding the dimensionality, convolving a 
multichannel input $\boldsymbol{\mathsfit{I}}$ with $k$ multichannel kernels 
$\boldsymbol{\mathsfit{K}}_{:,:,:,k}$ of equal 
dimensions results 
in a tensor of feature maps 
$\boldsymbol{\mathsfit{M}}$. The 
above function $dim(\cdot)$ can now be generalized to $n$ dimensions. Assume a 
padding $p$ and a stride $s$, then

\begin{align}\label{eq:size_of_generalized_convolutions}
	dim(\boldsymbol{\mathsfit{I}}) &\stackrel{\text{def}}{=} \{w, h, c\}\\
	dim(\boldsymbol{\mathsfit{K}}) &\stackrel{\text{def}}{=} \{f, f, c, k\}\\
	dim(\boldsymbol{\mathsfit{M}}) &= \{\lfloor \frac{w - 2p - f }{s} + 
	1 
	\rfloor, \lfloor \frac{h - 2p - f }{s} + 
	1 
	\rfloor, k\},
\end{align}

where $\lfloor x \rfloor$ is the floor function that rounds its argument to 
the nearest integer.

Correspondingly, the $2D-$convolution (Eq. \ref{eq:cross_correlation}) 
may be generalized to a $3D-$convolution in layer $l$ as expected.
A specific element $\mathsfit{M}_{i,j,k}$ of a determined output 
channel $k$ is 
calculated by  
1) sliding the three-dimensional kernel 
$\boldsymbol{\mathsfit{K}}_{:,:,:,k}$ with weights $w_{m,n,c}$ across the 
three-dimensional input and then 2) computing the tensor 
inner product $\boldsymbol{\mathsfit{K}}_{:,:,:,k} \cdot 
\boldsymbol{\mathsfit{A}}$ of 
the kernel and the three-dimensional tensor $\boldsymbol{\mathsfit{A}}$ of 
overlapped input elements. Stacking the elements of all channels that have been 
generated by sliding $k$ kernels produces the layer's output volume 
$\boldsymbol{\mathsfit{M}}^{l}$ with elements $\mathsfit{M}{\,}^{l}_{i,j,k}$. 
That is,

\begin{align}
	{\big( (\boldsymbol{\mathsfit{I}} * \boldsymbol{\mathsfit{K}})(i,j,k) 
		\big)}^l 
	&= \sum_{c} 
	\sum_{m} \sum_{n} x^{l - 1}_{i + m - 1,j + n - 1,c} \cdot w^l_{m,n,c,k} 
	\label{eq:cross_correlation_3d_1}\\
	z^l_{i,j,k} &= {\big( (\boldsymbol{\mathsfit{I}} * 
		\boldsymbol{\mathsfit{K}})(i,j,k) \big)}^l + b^l\\
	\mathsfit{M}{\,}^{l}_{i,j,k} &= 
	z^l_{i,j,k}.\label{eq:cross_correlation_3d_3}
\end{align}

%In DL frameworks like \cite{}

\subsection{Pooling Layer}\label{subsec:pooling}

Yet another prominent method to reduce the amount of parameters in the network 
is to use a \textbf{pooling} layer. Pooling outputs the result of computing 
a determined summary statistic in neighborhoods of activations of the previous 
layer.

Like convolution, the pooling function may be viewed as sliding a multi-channel 
window over a multi-dimensional input with equal amount of channels. Therefore, 
the 
pooling function may be considered as a filter of squared size $f$
that may be shifted by a stride $s$, acting on a possibly $p$ zero-padded input.
Unlike the 
convolution layer, the pooling layer does not have weights that must be 
updated by the learning algorithm. 

Additionally, pooling is applied on each channel independently. For an input 
volume $\boldsymbol{\mathsfit{M}}$ with dimensions $\{w, h, c\}$, pooling's 
output $\boldsymbol{\mathsfit{O}}$ dimensionality is

\begin{equation}\label{eq:size_of_generalized_pooling}
	dim(\boldsymbol{\mathsfit{O}}) = \{\lfloor \frac{w + 2p - f}{s} + 1 
	\rfloor, \lfloor \frac{h + 2p -f }{s} + 1 \rfloor, c\}.
\end{equation}

While the pooling function may calculate several summary statistics on any 
combination of dimensions, the most 
used is the operator that yields the maximum among the corresponding channel 
input 
elements $\boldsymbol{\mathsfit{M}}_{:,:,k}$ inside the sliding window 
$\max(\boldsymbol{X})$, termed \textbf{max 
	pooling}. That is, in one-based index form,

\begin{equation}\label{eq:generalized_pooling}
	\boldsymbol{\mathsfit{O}}_{i,j,k}^l = 
	\max_m \max_n \boldsymbol{\mathsfit{M}}^{l - 1}_{(i \cdot s - 1) + (m - 1), 
		(j \cdot s - 1) + (n - 1), k}.
\end{equation}

%Other forms of max pooling include pooling over the channel dimension 
%
%In Sec. regular. we discussed some regularizations techniques. This is also 
%one.

\subsection{ReLU layer}\label{subsec:relu}

As we mentioned, we adopted the simple layer terminology, where the 
non-linearities are not part of the convolutional layer. Instead, we consider 
that there is a \textbf{ReLU layer} which may be integrated into the network's 
architecture. As expected, the ReLU layer computes Eq. \ref{eq:relu}.

\subsection{CNN Training}\label{subsec:cnn_training}

\subsubsection{Forward and Back-propagation in CNNs}\label{subsec}

Since CNNs are deep feedforward network (Sec. \ref{sec:ann}), its weights may 
be 
initialized using the Kaiming method (Subsec. \ref{subsub:backprop}). CNNs may 
also 
be 
trained with gradient descent \ref{subsubsec:gradient_descent}. In particular, 
forward propagation (Eqs. 
\ref{eq:cross_correlation_3d_1}-\ref{eq:cross_correlation_3d_3}, \ref{eq:relu}, 
and 
\ref{eq:generalized_pooling}) must be 
performed to obtain the error at the output layer, followed by backprop.

As we discussed before (Subsec. \ref{subsub:backprop}) while backpropagating 
the error, 
the derivatives of the loss function w.r.t. 
the parameters need to be computed in order to update the weights. For CNNs, we 
will follow the strategy that we described in Subsec. \ref{subsub:backprop}. 
That is, 1) compute the error 
at the output layer, 2) computer the error of the current layer in terms of the 
next layer to be able to backpropagate, 3) compute the weight derivatives 
in terms of the error at the current layer and 4) compute the derivative of the 
bias.

In a CNN, the convolutional layer's parameters are the kernels' weights and the 
layer biases, one for each kernel. 
Therefore, the goal of backprop may be restated as computing the derivatives of 
the loss w.r.t. the kernels and the bias. Additionally, the derivatives of 
the ReLU and max pooling layer must be considered.

Since now we are operating with 3D tensors, we define the derivative of the 
cost function w.r.t one input element $z^l_{x,y,z}$ as

\begin{equation}
	\delta^l_{x,y,z} = \frac{\partial J}{\partial z^l_{x,y,z}}.
\end{equation}  

Commonly, the output layer of a CNN is a dense layer. Therefore, the error at 
the output 
layer $\boldsymbol{\delta}^\ell$ may be computed in a the same manner as for a 
general ANN (Eq.
\ref{eq:error_output_layer_matrix_form}). As we will see, the architecture of CNNs comprises ReLU and max pooling layers. 
Therefore, the computation of the backpropgation and concrete weights (Eqs. 
\ref{eq:final_layer_one_layer_minus_one_matrix} and 
\ref{eq:partial_with_respect_to_weight_3}) must be adapted to these layers.

Fortunately, neither the ReLU layer nor the max pooling layer contain weights 
that must be 
updated by SGD. That means Eqs. 
\ref{eq:partial_with_respect_to_weight_3}-\ref{eq:gradient_cost_wrt_bias} 
must 
be disregarded for these two layers. However, in our strategy these layers must 
be able to transmit 
the error back to preceding layers, which indeed requires readjusting Eq. 
\ref{eq:final_layer_one_layer_minus_one_matrix} (backpropagation equation 
between layer $l + 1$ and $l$). For 
that matter, we may consider that the error at these layers are 
known, e.g., after having been backpropagated by a dense layer.

On order to adapt Eq. \ref{eq:final_layer_one_layer_minus_one_matrix} to the 
ReLU layer, note that, because the layer does not contain weights, the weigh 
matrix 
$\boldsymbol{W}$ vanishes. Also, we now consider the local gradient 
$\delta^l_{i,j,k}$ to be a tensor element. Since forward propagating through a 
ReLU layer is conducted elementwise, the input and output dimensions are 
preserved, and so happens with backpropagation. This allows equal indexing of 
the backpropagated error arriving at the layer and the calculated local 
gradient. That is,

\begin{equation}\label{eq:relu_layer_backprop}
	\delta^l_{i,j,k} = \delta^{l + 1}_{i,j,k} \cdot	g^\prime(z^l_{i,j,k}),
\end{equation}
where, $g^\prime(z^l_{i,j,k})$ is the derivative of the ReLU evaluated at the 
tensor element $\boldsymbol{\mathsfit{M}}^{l}_{i,j,k}$.

Conversely, one key observation is that the max pooling layer does change the 
dimensionality of its input. Specifically, it  downsamples the input by 
discarding its activations in the sliding window, except for the
one with the maximum value. These units, whose 
activations were discarded, will obviously
receive zero gradient.

During forward propagation, the max pooling operation does more than just select the maximum value in each window; it also "memorizes" the indices of these maximum values within the tensor. This creates a one-to-one mapping between each pooled activation $\boldsymbol{\mathsfit{M}}_{m,n,k}$---the maximum value in each region---and the corresponding output element $\boldsymbol{\mathsfit{O}}_{i,j,k}$, based on these retained indices. When backpropagating, only the unit whose index has been memorized receives a non-zero gradient. This is called \textbf{gradient routing} because the 
max pooling layer routes back the gradient it receives to the unit whose maximum value was selected. For example, if the maximum value in a pooling window is at $\boldsymbol{\mathsfit{M}}_{1,2,3}$, and this value is mapped to an output element, say $\boldsymbol{\mathsfit{O}}_{0,1,3}$, the index $\{1,2,3\}$ is preserved.
Consequently, during backpropagation, this memorized index ensures that the gradient flows back to and only to $\boldsymbol{\mathsfit{M}}_{1,2,3}$.

Because the partial derivative of the max operation $\frac{\partial 
	g}{\partial{ 
		\boldsymbol{\mathsfit{M}}_{m,n,k}}}$ w.r.t the pooled value 
$\boldsymbol{\mathsfit{M}}_{m,n,k}$ equals one, Eq. 
\ref{eq:final_layer_one_layer_minus_one_matrix} takes the form 

\begin{equation}\label{eq:max_pooling_layer_backprop}
	\delta^l_{m,n,k} = 
	\begin{cases}
		\delta^{l + 1}_{i,j,k} & \text{if } \, \,			
		\boldsymbol{\mathsfit{M}}_{m,n,k} = 
		\boldsymbol{\mathsfit{O}}_{i,j,k}\\
		0 & \text{otherwise.}
	\end{cases}
\end{equation}

Having obtained the backpropagation rules of the ReLU and max pooling layer, it 
remains to discussed backpropagation on the convolutional layer. Indeed, in 
this case, the kernel weights and the units bias must be updated. Consequently, 
1) the backpropagation rule of the convolution layer and 2) the derivative of 
the loss function w.r.t. 
the kernel weights and neuron bias must be derived.

First, let us consider the backpropagation rule for the convolutional layer. 
Like for fully connected layers, we start by observing that the chain rule may 
be employed to compute the neuron error in layer $l$.  Also, to unclutter 
notation, we will assume zero padding and unit stride ($p = 0$, $s = 1$).

Unlike the general case,  
one input element in a convolutional layer $l$ affects one or more elements 
in layer $l + 1$, but not all. Therefore, the partial derivative of the cost 
function w.r.t. 
one element $z^l_{m,n,c}$ must aggregate the partial derivatives of all 
elements $z^{l + 1}_{\cdot,\cdot,\cdot}$ affected by it.

Because the 
convolutional layer 
downsamples the input spatial dimensions, 
careful must be taken when selecting the output element indices whose 
contributions must be added. One important 
observation is that one input element affects all feature maps (albeit not all 
elements), thus we must sum over all output channels $k$. Another observation 
is that non multi-channel kernel $\mathsf{K}_{m,n,c}$ move 
across the depth dimension, therefore, their index does not contain offsets. 
After having considered the 
channels, we must further consider the effect of the input element 
$z^l_{m,n,c}$ in one 
specific output channel $z^l_{:,:,k}$. Given a 
specific channels $k$, the input element affects a number of elements in that 
channel, namely the elements fixed at the position $m,n$ and that are the 
result of the kernel 
having been slit by by $-a, -b$ (input perspective, we introduce auxiliary 
indices $a$ and $b$). Finally, we know that all 
kernels have the same dimensionality (they are packed in one 4D-tensor), thus, 
we can use the same iterators $a,b$ for all kernels. These observations lead to

\begin{align}
	\delta^l_{m,n,u} &= \sum_{k} \sum_a \sum_b \frac{\partial 
		J}{\partial \; z^{l + 
			1}_{m - a + 1,n - b + 1,k}} 
	\cdot \frac{\partial z^{l + 1}_{m - a + 1,n - b + 1,k}}{\partial 
		z^l_{m,n,u}}\label{eq:cnn_backprop_rule_2}\\
	&= \sum_{k} \sum_a \sum_b \delta^{l + 
		1}_{m - a + 1,n - b + 1,k} 
	\cdot \frac{\partial z^{l + 1}_{m - a + 1,n - b + 1,k}}{\partial 
		z^l_{m,n,u}}.\label{eq:cnn_backprop_rule_3}
\end{align}

In Eq. \ref{eq:cnn_backprop_rule_3}, the neuron error $\frac{\partial 
	J}{\partial \; z^{l + 1}_{m - a + 1,n - b + 1,k}}$ has been 
substituted by its definition $\delta^{l + 
	1}_{m - a + 1,n - b + 1,k}$.

We now focus in the second term in the right and expand it. Here, again we 
introduce auxiliary indices $p, q, r$, and use $x^l_{\cdot,\cdot,\cdot} = 
z^l_{\cdot,\cdot,\cdot}$ (as 
mentioned, we assumed we are operating on linear layers).

\begin{align}
	\frac{\partial z^{l + 1}_{m - a + 1,n - b + 1,k}}{\partial z^l_{m,n,c}} &= 
	\frac{\partial}{\partial z^l_{m,n,c}} \Big( 
	\sum_{p} \sum_{q} \sum_{r} z^{l}_{(m - a + 1) + p - 1,(n - b + 1) + q - 
		1,r} 
	\cdot w^{l + 
		1}_{p,q,r,k} + b^l\Big)\label{eq:cnn_backprop_rule_4}\\
	&= 
	\frac{\partial}{\partial z^l_{m,n,c}} \Big( 
	\sum_{p} \sum_{q} \sum_{r} z^{l}_{m - a + p,n - b + q,r} 
	\cdot w^{l + 1}_{p,q,r,k} + b^l\Big)\label{eq:cnn_backprop_rule_5}\\
	&= 
	\frac{\partial}{\partial z^l_{m,n,c}} \Big( z^{l}_{m,n,c} 
	\cdot w^{l + 1}_{a,b,c,k}\Big)\label{eq:cnn_backprop_rule_6}\\
	&= w^{l + 1}_{a,b,c,k}.\label{eq:cnn_backprop_rule_7}
\end{align}

In Eq. \ref{eq:cnn_backprop_rule_4} we expanded as mentioned above, then in Eq. 
\ref{eq:cnn_backprop_rule_5}, we 
reduced the indices as $(m - a + 1) + p - 1 = m - a + p, (n - b + 1) + q - 1 = 
n - b + q$. In Eq. \ref{eq:cnn_backprop_rule_3}, it may be observed that all 
partial derivatives w.r.t. 
$z^l_{m,n,c}$, except when $a = p, b = q, r = c$. Accordingly, the weights must 
have indices $a,b,c,k$. Then, plugging Eq. \ref{eq:cnn_backprop_rule_7} in 
Eq.\ref{eq:cnn_backprop_rule_3}, we 
have:

\begin{align}
	\delta^l_{m,n,c} &= \sum_{k} \sum_a \sum_b \delta^{l + 
		1}_{m - a + 1,n - b + 1,k} 
	\cdot w^{l + 1}_{a,b,c,k}.\label{eq:cnn_backprop_rule_8}\\
	\delta^l_{m,n,c} &= \delta^{l +	1}_{m,n,c}  \; * \; 
	\text{rot}_{{180}^\circ} (w^{l + 
		1}_{a,b,c,k}).\label{eq:cnn_backprop_rule_9}
\end{align}

Eq. \ref{eq:cnn_backprop_rule_8} is a standard convolution of the error at the 
end of the 
convolutional layer (the error it receives from 
the following layer) with the kernel tensor. Equivalently, Eq. 
\ref{eq:cnn_backprop_rule_8} may expressed as a 
cross-correlation with flipped kernel (Eq. \ref{eq:cnn_backprop_rule_9}) by 
rotating 
the 3D kernel 
$180^\circ$ around the depth axis ($\text{rot}_{{180}^\circ}(\cdot)$). Note in 
Eq. 
\ref{eq:cnn_backprop_rule_8} that the matrix with elements 
$\delta^{l + 1}_{\cdot,\cdot,\cdot}$ has reduced spatial dimensions compared 
with the 
input $I$ to the convolutional layer. Therefore, to backpropagate, the matrix 
must be zero-padded (upsampled) by the amount $f - p - 1 = f - 1$ (recall $f$ 
is the filter 
size) such as after having been convolved, $I$'s 
dimensions are obtained. In another words, backpropagation requires 
a \textbf{transposed convolution}. Furthermore, Eq. 
\ref{eq:cnn_backprop_rule_8} suggests that, in order to propagate the error to 
the input unit $\delta^{l}_{m,n,c}$, one must collect the kernel weights in the 
channel $c$ of all kernels $k$, and multiply them with the corresponding 
elements in the gradient tensor at the corresponding $k$ feature map.

An interesting concrete example in given by \cite{Aggarwal2018}:

``In order to understand the transposition above, consider a situation in which 
we use
20 filters on the 3-channel RGB volume in order to create an output volume of 
depth 20.
While backpropagating, we will need to take a gradient volume of depth 20 and 
transform
to a gradient volume of depth 3. Therefore, we need to create 3 filters for 
backpropagation,
each of which is for the red, green, and blue colors. We pull out the 20 
spatial slices from
the 20 filters that are applied to the red color, invert them (...),
and then create a single 20-depth filter for backpropagating gradients with 
respect to the
red slice. Similar approaches are used for the green and blue slices.´´

Similarly, the derivative of the cost function w.r.t. to the kernel weights 
$w^l_{\cdot,\cdot,\cdot}$ must be computed. A kernel weight $w^l_{m,n,c,k}$ 
affects only the elements in the $k$ feature map. Therefore, when applying the 
chain rule, its gradient must add the 
partial derivative of all those elements:

\begin{align}
	\diffp[]{J}{{w^l_{m,n,c,k}}} &= \sum_i \sum_j \diffp[]{J}{{z^{l}_{i,j,k}}}
	\cdot 
	\diffp[]{{z^{l}_{i,j,k}}}{{w^l_{m,n,c,k}}}\label{eq:cnn_backprop_rule_11}\\
	&= \sum_i \sum_j \delta^{l}_{i,j,k}
	\cdot 
	\diffp[]{{z^{l}_{i,j,k}}}{{w^l_{m,n,c,k}}}.\label{eq:cnn_backprop_rule_12}
\end{align}

The same techniques as before may be applied in Eq. 
\ref{eq:cnn_backprop_rule_11}, i.e., substitute the neuron error by its 
definition which results in Eq. \ref{eq:cnn_backprop_rule_12}. Then the second 
term in the right may be expanded as 

\begin{align}
	\diffp[]{{z^l_{i,j,k}}}{{w^l_{m,n,c,k}}} &= \diffp[]{{}}{{w^{l}_{m,n,c,k}}} 
	\Big( 
	\sum_{r} \sum_{p} \sum_{q}  z^{l}_{i  +  p - 1,j + q - 1,r} \cdot 
	w^{l}_{p,q,r,k}  + 
	b^l \Big)\label{eq:cnn_backprop_rule_10}\\
	&= \diffp[]{{(z^{l}_{i + m - 1,j + n -1,c} \cdot 
			w^{l}_{m,n,c,k})}}{{w^l_{m,n,c,k}}}\label{eq:cnn_backprop_rule_13}\\
	&= z^{l - 1}_{i + m - 1, j + n - 1,c}. \label{eq:cnn_backprop_rule_14}
\end{align}

Calculating the partial derivative of the triple summation in 
Eq.\ref{eq:cnn_backprop_rule_10} reduce all terms to zero except the term with 
index $p = m$, $q = n$ and $r = c$. The result (\ref{eq:cnn_backprop_rule_14}) 
may be substitute in Eq. \ref{eq:cnn_backprop_rule_12}, i.e.,

\begin{align}
	\diffp[]{J}{{w^l_{m,n,c,k}}} &= \sum_i \sum_j \delta^{l}_{i,j,k}
	\cdot 
	z^{l - 1}_{i + m - 1, j + n - 1,c}\label{eq:cnn_backprop_rule_15}\\
\end{align}

Comparing Eq. \ref{eq:cnn_backprop_rule_13} to Eq. 
\ref{eq:cnn_backprop_rule_9}, it may be noted that while backpropagating the 
error both the kernel and the layer error must be rotated.

The derivation of the error of the cost function w.r.t. the bias 
follows also the above strategy. The shared bias affects all activations in the 
feature map. However, from Eq. \ref{eq:cross_correlation} may be concluded that 
the partial derivative ot all activations $i,j,k$ w.r.t. the bias equal one, 
i.e., $\frac{\partial z^l_{i,j,k}}{\partial b^l} = 1$. Thus,

\begin{align}
	\diffp[]{J}{{b^l}} &= \sum_k \sum_a \sum_b  \diffp[]{J}{{z^l_{a,b,k}}}
	\cdot \diffp[]{{z^l_{a,b,k}}}{{b^l}}\label{eq:cnn_backprop_rule_16}\\
	&= \sum_k \sum_a \sum_b \delta^l_{a,b,k}.\label{eq:cnn:backprop_15}
\end{align}

\subsection{Batch Normalization}\label{subsec:batch_norm}

One of the earliest findings in machine learning was that imputing features in 
different scales hinders ANN learning \cite{bishop1995neural} \cite{LeCun2012}.
Except for settings where a preprocessing of the features is not desired, 
feature normalization is standard practice.

While features are supplied to the ANN at the input layer, the question arises 
naturally if the same strategy may be applied to the hidden layers. As shown by 
Eq. \ref{eq:ann_composition}, a determined hidden layer's input is a highly 
non-linear function of 
the ANN inputs. From the 
perspective of the hidden layer, the input to its units constantly varies in 
possibly several scales, complicating learning.

This motivates the \textbf{Batch normalization} (also 
\textbf{Batchnorm}) method \cite{batchnorm_2015}. Batchnorm proposes to 
normalize 
the input to a layer 
to stabilize optimization. This method may be presented in several forms (cf. 
\cite{batchnorm_2015}, \cite{Goodfellow.2016}). Specifically, while the former 
explains Batchnorm from a transform perspective, the latter considers it  
a reparametrizing technique: `` Batch normalization provides an elegant way of 
reparametrizing almost any deep network". Here, we follow this approach.

Consider a mini-batch of
activations (output) at a 
determined layer $l - 1$, which may be normalized as to train the layer $l$ 
faster. This may be 
represented by a design matrix 
$\boldsymbol{H} 
\in \mathbb{R}^{n \times 
	m}$ 
with elements $a_{ij}$,
where $i$ is the activation of unit $i$ corresponding to one mini-batch 
example $x^{(j)}$. That 
is, the rows of $\boldsymbol{H}$ are the activations vector corresponding to 
one 
training 
example, while the columns are the activations at unit $i$ of all mini-batch 
training examples.

Batchnorm normalizes each activation independently by subtracting the mean 
$\mu$ and 
dividing by the 
standard deviation $\sigma$ in a 
per-column basis. That is,

\begin{align}
	\mu_j &= \frac{1}{m} \sum_{i = 1}^{m} a_{ij} \label{eq:batchnorm_mu}\\
	\sigma_j &= \delta + \frac{1}{m} \sqrt{\sum_{i = 1}^{m} | a_{ij} - \mu_j 
		|^2} 
	\label{eq:batchnorm_sigma}\\
	\boldsymbol{\mu} &= \{ \mu_1, \cdots, \mu_m\} \label{eq:batchnorm_mu_vector}\\
	\boldsymbol{\sigma} &= \{ \sigma_1, \cdots, \sigma_m\} 
	\label{eq:batchnorm_sigma_vector}\\
	\boldsymbol{H}^\prime &= \frac{\boldsymbol{H} - 
		\boldsymbol{\mu}}{\boldsymbol{\sigma}},\label{eq:batchnorm_normalization}
\end{align}

where the components of the mean vector $\boldsymbol{\mu}$ (Eq. 
\ref{eq:batchnorm_mu_vector}) and the 
standard deviation 
vector $\boldsymbol{\sigma}$ (Eq. \ref{eq:batchnorm_sigma_vector}) are the 
mean (Eq. 
\ref{eq:batchnorm_mu}) and the standard 
deviation (Eq. \ref{eq:batchnorm_sigma}) of each unit. In Eq. 
\ref{eq:batchnorm_sigma}, a small 
constant $\delta$ (e.g., $\mathnormal{10^{-8}}$) is added for numerical 
stability. In Eq.
\ref{eq:batchnorm_normalization} the vectors $\boldsymbol{\mu}$ and 
$\boldsymbol{\sigma}$ are broadcasted to normalize $a_{ij}$ using $\sigma_{j}$ and 
$\mu_{j}$.

Albeit intuitive, restricting the distribution of the activation to have zero 
mean and unit standard deviation is an arbitrary choice. \textit{A 
	priori}, it is not guaranteed that this setting will result in accelerating 
learning. Optimally, the ANN should learn how to transform the activations.

Indeed, instead of replacing $\boldsymbol{H}$ by its normalized form 
$\boldsymbol{H}^\prime$, the activations may be replaced with 
the parametrized form $BN(\boldsymbol{H}) = \boldsymbol{\gamma} 
\boldsymbol{H}^\prime + 
\boldsymbol{\beta}$, where the vectors $\boldsymbol{\gamma}$ and 
$\boldsymbol{\beta}$ may 
be learned by SGD, say. The arithmetic here is as described above, i.e., the 
activation $a_{ij}$ is replaced by

\begin{equation}
	\gamma_j \cdot \frac{a_{ij} - {\mu_j}}{\delta_j} + \beta_j.
\end{equation}

This form has the advantage that it can recover the 
original normalized activations $\boldsymbol{H}^\prime$ for 
$\boldsymbol{\gamma} = \boldsymbol{\mathnormal{1}}, \boldsymbol{\beta} = 
\boldsymbol{\mathnormal{0}}$, and, more importantly allows the activation 
distribution to be arbitrarily shifted and scaled by the learning algorithm.

During training, the learning algorithm updates the parameters 
$\boldsymbol{\beta}$ and $\boldsymbol{\delta}$ so that when training has finished, 
these 
parameters are ready to use for inference. In contrast, the empirical 
statistics $\mu$ and $\delta$ used to normalize the activations are computed on a 
mini-batch (Eqs. \ref{eq:batchnorm_mu}-\ref{eq:batchnorm_sigma}). This may pose 
two problems. First, during inference with one instance, the statistics are not 
well defined (because of the unit batch size). Second, if inference is  
conducted for a mini-batch, the relation of one determined instance with respect 
to its predicted target is not deterministic because the statistics depend on the 
batch in which this instance may appear.

One standard solution (that we also use) is to maintain running averages of the 
statistics during training, i.e., for each activation, $\mu$ and $\sigma$ are 
averaged 
over all mini-batches in the training set. Then, inference is conducted with these 
fixed statistics.

The effect of Batchnorm is to some extent controversial in the scientific 
community. 
For example, while the authors of the original paper claimed that Batchnorm 
diminish the constant variation of the mean and variance of the hidden layers 
(internal covariance shift) \cite{batchnorm_2015}, other scientists have shown
that Batchnorm's efficiency is based on a different mechanism 
\cite{how_batchnorm_works_2018}. That being said, Batchnorm is an indispensable part of ANNs.

\section{Summary}\label{sec:summary}

In this tutorial we examined the foundations of Deep Learning with special focus on Convolutional Neural Network. Deep Learning belongs to a wider set of methods called Machine Learning, which main goal is to develop system or agents that are able to learn from data.

The most important concept of learning is generalization. That is, given a 
dataset, the learning algorithm is trained on the data to perform a specific task, and is required to accomplish the task on data it did not experience before. In particular, is expected that the agent outputs a certain type of inference or prediction based on the data. Common tasks are classification (predicting a category) and regression (predicting a continuous quantity), for example, classifying images of cars and regressing the height of person in a picture. 

The dataset is a key component of the learning algorithm. It is composed of examples or observations, each having features that contain useful information for training the agent. The dataset may include a prediction target for every example, such as the agent can learn by processing the examples and observing the correct prediction it has to output. This is called supervised learning and the targets accompanying the examples are the ground truth. The opposite case, in which the dataset does not contain ground truth is named unsupervised learning. Additionally, there other two forms of learning, namely, semi-supervised learning and reinforcement leaning, but this tutorial focused on supervised learning.

Deep learning borrows concepts from Learning theory, and so it assumes that there is function in a hypothesis space that maps the examples to their targets. In this view, the learning agent is the model that approximates the function in the hypothesis space. The less generalization error, the better the approximation. Theoretically, the dataset may be divided in a training set, where the training error may be calculated, and a test set where generalization error is supposed to be assessed. However, while training, the agent has no access to the test set and can only estimate the generalization error using the training set. This error is called empirical risk, therefore, the agent general strategy from the perspective of leaning theory is termed Empirical Risk Minimization. 

In addition to Learning Theory, Deep Learning also borrows concepts from Statistics and Probability theory. In particular, the leaning agent is considered an estimator that may be parameterized. When analyzing ANNs, these parameters are conceptualized as weights. This leads to the conclusion that the estimator function $is$ the neural network.

Artificial Neural Networks are comprised of artificial neurons. The neurons 
compute a non-linear function of the linear combination of the inputs.

To train the network, gradient descent may be used. Gradient descent requires 
updating the network weights. In order to compute the 
suitable update magnitude, three steps must be iteratively conducted. First, 
forward propagate 
the information from the input layer, through the neurons, to the output layer. 
Second, calculate the error of the cost function w.r.t. parameters at the output 
layer, and finally, backpropagate the error until the input layer.

Most of the fundamental principles in which DL rest have been developed in the 
decades ago or even before. For example, statistical learning, 
learning theory, SGD, Perceptron, Backprop, CNN. However they remain being used 
in all modern DL applications.

Several of the most important innovations in the DL were inspired by the 
structure and functioning of human and animal brain. Perceptron, ReLU, 
receptive fields (CNN), BatchNorm.

Although regression tasks may be cast to classification tasks by discretizing the output, regression has a number of advantages compared to classification when the task involves predicting continuous quantities. For example regression provides higher resolution, less information loss, or, depending on the manner the problem is framed, more appropriate error metrics. In cases like weather prediction (temperature and future rainfall) regression is optimal.

While regression offers precise predictions and is suitable for continuous data, it comes with challenges such as sensitivity to outliers, complexity in evaluating performance, and difficulty in handling nonlinear relationships or imbalanced data. For some problems, these factors can make classification a more straightforward or effective approach.

That being said, deep regression is a powerful framework that can be employed to tackle several problems in the real world.

%%%%%%%%%%%%%%%%%%%%%%%%%%%%%%%%%%%%%%%%%%
\vspace{6pt} 

\bibliographystyle{apalike}  % or use another style like unsrt, alpha, etc.
\bibliography{./references}
\pagebreak
%!TEX root = ../dissertation.tex
% Notation

\textbf{Notation}\\
In this tutorial, we use the following notation:
%{\def\arraystretch{1.5}\tabcolsep=8pt
\renewcommand*{\arraystretch}{1.6}
\begin{longtable}{lp{10cm}}%[h!]
	%\begin{tabular}{cl}
		\multicolumn{2}{l}{\textbf{Number and Arrays}}\\
		$a$ & A scalar (integer or real)\\
		$\boldsymbol{a}$ & A vector\\
		$\boldsymbol{A}$ & A matrix\\
		$\boldsymbol{\mathsfit{A}}$ & A tensor\\
		$\textnormal{a}$ & A scalar random variable\\
		$\mathbf{a}$ & A vector-value random variable\\
		$\mathbf{A}$ & A matrix-value random variable\\
		\multicolumn{2}{l}{\textbf{Sets}}\\
		$\mathbb{A}$ & A set\\
		$\mathbb{R}$ & The set of real numbers\\
		$\{0, 1, \cdots, n\}$ & The set of all integers between $0$ and $n$\\
		\multicolumn{2}{l}{\textbf{Indexing}}\\
		$a_i$ & Element $i$ of vector $\boldsymbol{a}$\\
		$\boldsymbol{a}_{-i}$ & All elements of vector $\boldsymbol{a}$ except for 
		element $i$\\
		$\boldsymbol{A}_{i,j}$ & Element $i,j$ of matrix 
		$\boldsymbol{A}$\\
		$\boldsymbol{A}_{i,:}$ & Row $i$ of matrix $\boldsymbol{A}$\\
		$\boldsymbol{A}_{:,i}$ & Column $i$ of matrix $\boldsymbol{A}$\\
		$\mathsfit{A}_{i,j,k}$ & Element $(i,j,k)$ of a tensor 
		$\boldsymbol{\mathsfit{A}}$\\
		$\boldsymbol{\mathsfit{A}}_{:,:,i}$ & 2-D slice of a 3-D tensor\\
		$\boldsymbol{\mathsfit{A}}_{:,:,:,i}$ & 3-D slice of a 4-D tensor\\
		\multicolumn{2}{l}{\textbf{Linear Algebra Operations}}\\
		$\boldsymbol{A}^T$ & Transpose of matrix $\boldsymbol{A}$\\
		$\boldsymbol{A} \odot \boldsymbol{B}$ & Element-wise (Hadamard) product of 
		$\boldsymbol{A}$ and $\boldsymbol{B}$\\
		\multicolumn{2}{l}{\textbf{Calculus}}\\
		$\frac{dy}{dx}$ & Derivative of $y$ with respect to $x$\\
		$\frac{\partial{z}}{\partial{y}}$ & Partial derivative of $y$ with respect 
		to $x$\\
		$\nabla_{\boldsymbol{x}}y$ & Gradient of $y$ with respect to $x$\\
		\multicolumn{2}{l}{\textbf{Probability and Information Theory}}\\
		$P(a)$ & A probability distribution over a discrete variable\\
		$p(a)$ & A probability distribution over a continuous variable, or over a 
		variable whose type has not been defined\\
		$a \sim P$ & Random variable $a$ has distribution $P$\\
		$\mathbb{E}_{\mathbf{x} \sim P}[f(x)]$ or $\mathbb{E}f(x)$ & Expectation 
		of $f(x)$ 
		with respect to $P(\mathbf{x})$\\
		$D_{KL}(P \| Q)$ & Kullback-Leibler divergence of $P$ and $Q$\\
		$\mathcal{N}(\boldsymbol{x}; \boldsymbol{\mu}, \boldsymbol{\Sigma})$ & 
		Gaussian distribution over $\boldsymbol{x}$ with mean $\boldsymbol{\mu}$ 
		and covariance $\boldsymbol{\Sigma}$\\
		\multicolumn{2}{l}{\textbf{Functions}}\\
		$f: \mathbb{A} \rightarrow \mathbb{B}$ & Function $f$ with domain 
		$\mathbb{A}$ and range $\mathbb{B}$\\
		$f(\cdot)^{(i)}$ & function $i$ of an ordered set of functions\\
		$f \circ g$ & Composition of the functions $f$ and $g$\\
		$f(\boldsymbol{x}; \boldsymbol{\theta})$ & A function of 
		$\boldsymbol{x}$ 
		parametrized by $\boldsymbol{\theta}$\\
		$\log x$ & Natural logarithm of $x$\\
		$\sigma(x)$ & Logistic sigmoid, $\frac{1}{1 \; + \; \exp(-z)}$\\
		$\zeta(x)$ & Softplus, $\log (1 + \exp(z))$\\
		${\Vert \boldsymbol{x} \Vert}_p$ & $L^p$ norm of $\boldsymbol{x}$\\
		${\Vert \boldsymbol{x} \Vert}$ & $L^2$ norm of $\boldsymbol{x}$\\
		\multicolumn{2}{l}{\textbf{Datasets and Distributions}}\\
		$p_{data}$ & The data generating distribution\\
		$\hat{p}_{data}$ & The empirical distribution defined by the training set\\
		$\mathbb{X}$ & A set of training examples\\
		$\boldsymbol{x}^{(i)}$ & The $i$-th example (input) from a dataset\\
		$\boldsymbol{y}_i$, $y^{(i)}$, or $\boldsymbol{y}^{(i)}$ & The target 
		associated 
		with $\boldsymbol{x}^{(i)}$ for supervised learning\\
		$\boldsymbol{X}$ & The $m \times n$ matrix with input example 
		$\boldsymbol{x}^{(i)}$ in row $\boldsymbol{X}_{i,:}$ 
	%\end{tabular}
\end{longtable}
%}

\end{document}